Article

# CORE-ReID V2: Advancing the Domain Adaptation for Object Re-Identification with Optimized Training and Ensemble Fusion

Trinh Quoc Nguyen [1,2,*], Oky Dicky Ardiansyah Prima [1], Syahid Al Irfan [1,2], Hindriyanto Dwi Purnomo [3], and Radius Tanone [3]

[1] Graduate School of Software and Information Science, Iwate Prefectural University, Takizawa-shi 020-0693, Iwate, Japan; g236v201@s.iwate-pu.ac.jp (T.Q.N.); prima@iwate-pu.ac.jp (O.D.A.P.); g231v012@s.iwate-pu.ac.jp (S.A.I.)

[2] CyberCore Co., Ltd., Morioka-shi 020-0045, Iwate, Japan; trinh@cybercore.co.jp (T.Q.N.) ; syahid.irfan@cybercore.co.jp (S.A.I.)

[2] Department of Information Technology, Satya Wacana Christian University, Salatiga, 50711, Indonesia; hindriyanto.purnomo@staff.uksw.edu (H.D.P.); radius.tanone@uksw.edu (R.T.)

* Correspondence: g236v201@s.iwate-pu.ac.jp

**Abstract:** This study presents CORE-ReID V2, an enhanced framework building upon CORE-ReID. The new framework extends its predecessor by addressing Unsupervised Domain Adaptation (UDA) challenges in Person ReID and Vehicle ReID, with further applicability to Object ReID. During pre-training, CycleGAN is employed to synthesize diverse data, bridging image characteristic gaps across different domains. In the fine-tuning, an advanced ensemble fusion mechanism, consisting of the Efficient Channel Attention Block (ECAB) and the Simplified Efficient Channel Attention Block (SECAB), enhances both local and global feature representations while reducing ambiguity in pseudo-labels for target samples. Experimental results on widely used UDA Person ReID and Vehicle ReID datasets demonstrate that the proposed framework outperforms state-of-the-art methods, achieving top performance in Mean Average Precision (mAP) and Rank-k Accuracy (Top-1, Top-5, Top-10). Moreover, the framework supports lightweight backbones such as ResNet18 and ResNet34, ensuring both scalability and efficiency. Our work not only pushes the boundaries of UDA-based Object ReID but also provides a solid foundation for further research and advancements in this domain. Our codes and models are available at https://github.com/TrinhQuocNguyen/CORE-ReID-V2.

## 1. Introduction

Object Re-identification (ReID) focuses on retrieving specific object instances across diverse viewpoints [1-5] and has gained significant attention within the computer vision community due to its broad range of practical applications. Substantial advancements have been made in both supervised [6-9] and unsupervised ReID tasks [5,10,11], with most approaches employing backbone models originally designed for generic image classification tasks [12,13].

Unsupervised domain adaptation (UDA) for object ReID aims to transfer knowledge learned from a labeled source domain to accurately measure inter-instance affinities in an unlabeled target domain. Typical ReID tasks, such as person ReID and vehicle ReID, involve source and target domain datasets that do not share identical class identities. State-of-the-art UDA methods [11,14-18] generally adopt a two-stage training paradigm: (1) supervised pre-training on the source domain and (2) unsupervised fine-tuning on the target domain. During the second stage, pseudo-labeling strategies have demonstrated effectiveness in recent works [11,16,17]. These strategies iteratively alternate between generating pseudo-class labels through clustering target-domain instances and refining the network by training on these pseudo-classes. This iterative process helps the pre-trained source-domain model gradually capture inter-sample relationships within the target domain, even in the presence of label noise. Compared to fully supervised methods, which rely on large amounts of labeled target data for optimal performance, UDA methods are more scalable and cost-effective but may suffer from reduced label reliability and slower convergence. The key advantage of supervised approaches lies in their access to precise annotations, while the strength of unsupervised methods is their ability to generalize to new domains without manual labeling, making them particularly valuable in real-world Re-ID deployments.

The integration of global features, which encapsulate coarse semantic information, and local features, which provide fine-grained details, has proven effective in enhancing algorithm performance. CORE-ReID [11] introduced the Ensemble Fusion framework which combines global and local features with the Efficient Channel Attention Block (ECAB). ECAB leverages inter-channel relationships to guide the model's attention toward salient structures within the input image. Although CORE-ReID achieves competitive results in Person ReID under the UDA setting, it has three main limitations. Firstly, ECAB is exclusively to local features, leaving global features unenhanced. Secondly, CORE-ReID only supports deep and complex backbone networks, such as ResNet50, ResNet101, and ResNet152, while neglecting shallower architectures like ResNet18 and ResNet34, which offer computational efficient and are well-suited for resource-constrained environments. Finally, CORE-ReID is limited to the Person ReID task, restricting its applicability to other ReID scenarios.

In this paper, we present CORE-ReID V2, an enhanced version of CORE-ReID that addresses its limitations and introduces several novel contributions. CORE-ReID V2 not only achieves superior performance in UDA Person ReID but also extends its applicability to Object ReID tasks and supports lightweight backbone networks, such as ResNet18 and ResNet34, making it suitable for real-time systems and mobile devices. Building on the principles of LF2 [17] and CORE-ReID, we design a mean-teacher-based framework that iteratively learns multi-view features and refines noisy pseudo-labels through multiple clustering steps. We introduce the Ensemble Fusion++ module in CORE-ReID V2, which adaptively enhances both local and global features. This module applies ECAB to local features and the Simplified Efficient Channel Attention Block (SECAB) to global features, resulting in a fused representation that provides a more comprehensive feature set. Furthermore, we improve clustering outcomes by incorporating the KMeans++ [19] initialization strategy, which balances randomness and centroid selection to enhance cluster quality. To validate the framework, we pre-train the model on a source domain that integrates camera-aware style-transferred data for Person ReID and domain-aware style-transferred data for Vehicle ReID. Additionally, we adopt a teacher-student architecture for iterative domain adaptation, which has been used in prior works such as LF2 [17], Deep Mutual Learning (DML) [20], MMT [15], MEB-Net [21] and Mean Teacher [22]. In our framework, the teacher network captures global features while the student network refines diverse local features, both contributing to a more effective pseudo-labeling process. To summarize, the key contributions of CORE-ReID V2 are as follows:

- **Advanced Data Augmentation Techniques:** The framework integrates novel data augmentation strategies, such as Local Grayscale Patch Replacement and Random Image-to-Grayscale Conversion for UDA task. These methods introduce diversity in the training data, enhancing the model's stability.
- **Dynamic and Flexible Backbone Support:** CORE-ReID V2 extends compatibility to smaller backbone architectures, including ResNet18 and ResNet34, without compromising performance. This flexibility allows for deployment in resource-constrained environments while maintaining high accuracy.

- **Expansion to Vehicle and further Object ReID:** Unlike its predecessor, which focused solely on person re-identification, CORE-ReID V2 extends its scope to Vehicle Re-identification and further general Object Re-identification. This expansion demonstrates its versatility and adaptability across various domains.
- **Introduction of Ensemble Fusion++:** The framework incorporates the SECAB into the global feature extraction pipeline to enhance feature representation by dynamically emphasizing informative channels, thereby improving discrimination between instances.

## 2. Related Work

Extensive research has been conducted on UDA for Object ReID [23-27] and knowledge transfer techniques, such as knowledge distillation, which enable well-trained models to transfer expertise and improve learning complex domain scenarios [28-31]. Methods generally fall into two categories: domain translation [32-41], which aligns visual styles between domains, and pseudo-labeling [11,14-17,42-45], which iteratively cluster target samples to generate pseudo-labels. While pseudo-labeling has demonstrated superior performance, both methods face challenges related to domain shifts and noisy labels. Moreover, the fusion of global and local features has proven effective in various tasks, including classification [46-51], object detection [52-57], semantic segmentation [58-63], by integrating contextual information with fine-grained details. Building on these insights, our work refines the CORE-ReID [11] fusion module to achieve a better balance between global and local features, leading to improved performance across multiple ReID tasks, including Person and Vehicle ReID.

*2.1. UDA for Object ReID*

UDA has gained significant attention for its ability to reduce reliance on costly manual annotations. By utilizing labeled data from a source domain, UDA enhances model performance in a target domain without requiring target-specific annotations. Research in Object ReID has primarily concentrated on Person ReID and Vehicle ReID [64]. Existing UDA approaches for ReID can be broadly grouped into two categories: domain translation-based methods and pseudo-label-based methods [65,66].

**Domain translation-based methods:** These methods align the visual style of labeled source domain images with that of the target domain. The translated images, along with their original ground-truth labels, are then used for training [37].

Several methods attempt to map source and target distributions to mitigate domain shifts [32-36]. Saenko et al. [32] introduced a domain adaptation technique based on cross-domain transformations by learning a regularized non-linear transformation that brings source domain points closer to the target domain. In [33], the Geodesic Flow Kernel (GFK) was proposed to address domain shifts by integrating an infinite number of subspaces that capture geometric and statistical changes between the source and target domains. Similarly, Fernando et al. [34] developed a mapping function to align the source subspace with the target subspace for improved adaptation. Correlation Alignment (CORAL) [35] addressed domain shifts by computing the covariance statistics of each domain and applying a whitening and re-coloring linear transformation to align the source feature with the target domain. The Disentanglement Then Reconstruction (DTR) framework [36] enhanced alignment by disentangling the distributions and reconstructing them to ensure consistency across domains.

Another line of research [38-41] adopts adversarial approaches to learn transformations in the pixel space between domains. Methods like PixelDA [38], PTGAN [39], and SBSGAN [40] enforce pixel-level constraints to preserve color consistency during domain translation. CoGAN [41] extends this concept by learning joint distributions, such as the joint distribution of color and depth images or face images with varying attributes.

Other methods focus on discover a domain-invariant feature space to bridge domain gaps [66-72]. SPGAN [66] and CGAN-TM [69] improve feature-level similarity between translated and original images. Deep Adaption Network (DAN) [71] employs the Maximum Mean Discrepancy (MMD) [73-75] to align feature distributions across domains. Similarly, Ganin et al. [67] and Ajakan et al. [76] introduced a domain confusion loss to encourage the learning of domain-invariant features. Hoffman et al. [70] proposed the Intermediate Domain Module (IDM) to generate intermediate domain representations dynamically by mixing the hidden features of the source and target domains through two domain features. CyCADA [72] combines both pixel-level and feature-level adaptation to improve domain adaptation.

**Pseudo-label-based methods:** The second category, pseudo-labeling methods [11,14-17,42-45], models the relationships between unlabeled target-domain data with generated pseudo labels. Fan et al. [42] proposed the progressive unsupervised learning (PUL) method that alternates between assigning labels to unlabeled samples and optimizing the network using the generated targets. This iterative refinement aligns the model's representations more closely with the target domain, enhancing adaptation over time. Lin et al. [43] developed a bottom-up clustering framework enhanced by a repelled loss mechanism, which aims to increase the discriminative power of learned features while mitigating intra-cluster variations. Similarly, UDAP [14] proposed a self-training scheme that minimizes loss functions iteratively using clustering-based pseudo labels. SSG [16], $LF^2$ [17], and CORE-ReID [11] further contributed to this category, which introduce techniques to assign pseudo labels to both global and local features. Ge et al. [15] proposed Mutual Mean Teaching (MMT), which combines offline hard pseudo labels and online soft pseudo labels in an alternating training process, enhancing the model's ability to adapt to domain shifts. This technique improves the model's capacity to handle domain shifts by iteratively refining both the pseudo labels and feature representations throughout training. SpCL [44] advanced this field by using a hybrid memory module that stored centroids of labeled source domain images alongside un-clustered target instances and target domain clusters. This hybrid memory provides additional supervision to the feature extractor, while minimizing a unified contrastive loss over the three types of stored information. Additionally, Zheng et al. [45] developed the Uncertainty-Guided Noise Resilient Network (UNRN), which evaluates the reliability of predicted pseudo labels for target domain samples. By incorporating uncertainty estimates into the training process, UNRN improves performance with noisy annotations, thereby enhancing performance in domain adaptation scenarios.

Pseudo-labeling methods analyze data at different levels of detail, allowing them to capture small differences within the target domain. While these methods have demonstrated strong performance in many recent UDA ReID studies [11,15,44,45], their effectiveness can vary depending on factors such as dataset bias, domain shift severity, clustering quality, and label noise. Domain translation methods, on the other hand, remain valuable for reducing style mismatches and have shown advantages in specific scenarios, especially when high-quality translated images can be generated [77-79].

*2.2. Knowledge Transfer*

Knowledge transfer is a broad concept that refers to using knowledge gained from one model, dataset, or task to improve performance in another [28-31]. Within this broad scope, knowledge distillation is a specific technique in which a "teacher" model guides a "student" model by transferring soft predictions, features, or intermediate representations [80]. Knowledge distillation techniques help student networks become more accurate and generalize better, as the teacher model's output implicitly contains rich information about the relationships between training samples and their underlying distribution [81]. For example, Laine and Aila [82] introduced the Mean Teacher model which averaged model weights across multiple training iterations to guide supervision for unlabeled data. In contrast, Deep Mutual Learning (DML) [20], proposed by Zhang et al., shifts from the traditional teacher-student framework by employing a group of student models that train collaboratively, providing mutual supervision and facilitating the exploration of diverse feature representations. Ge et al. introduced MMT [15], which adopts an alternative training method that uses both offline refined hard pseudo-labels and online refined soft pseudo-labels. MEB-Net [21] further builds on this by using three networks (six models in total) to conduct mutual mean teacher training and generate pseudo-labels.

In the context of our method, we adopt a teacher-student architecture aligned with the Mean Teacher framework [22], where the teacher network is updated through Exponential Moving Average (EMA) of the student's weights. This form of distillation encourages consistency and stability in the learned representations, particularly useful in UDA where target-domain labels are absent. For clarity, in this work, a domain refers to a specific dataset distribution (e.g., Market-1501 [83] or VehicleID [84]), typically captured under different environmental or camera conditions, while a task refers to the objective of re-identifying objects (e.g., people or vehicles) across domains with different identities and styles. Although the task remains constant (object ReID), our goal is to transfer the knowledge learned from a labeled source domain to an unlabeled target domain under domain shift conditions.

*2.3. Feature fusion*

The feature fusion of global and local features has proven highly effective across various computer vision tasks, including classification [46-51], object detection [52-57], semantic segmentation [58-63], and more [85].

In image classification, global features capture the overall structure and appearance, while local features focus on fine-grained details. Combining both types provides complementary information, enhancing the model's ability to generalize across variations such as pose, lighting, and occlusions. For example, an early approach to feature fusion based on Canonical Correlation Analysis (CCA) was proposed by Sun et al. [46] who applied CCA to extract correlation features between two groups of feature vectors for improved pattern recognition performance. This approach effectively captured discriminative information while reducing feature redundancy, demonstrating significant improvements in recognition rates on datasets such as CENPARMI and Yale Face Database compared to single-feature and traditional fusion methods. Later, Sudha and Ramakrishna [47] studied iris feature fusion with pixel-level methods like DU-Fusion and showed that combining features from techniques such as 2D-FFT, LBP, and PCA enhances recognition performance. The results on the CASIA dataset confirmed that DU-Fusion outperformed other methods in both verification and identification accuracy. Tian et al. [48] proposed a vehicle model recognition system using an iterative discrimination CNN based on selective multi-convolutional region feature extraction. Their SMCR model combines global and local features to boost classification accuracy. Similarly, Lu et al. [49] introduced a script identification framework that leverages both global CNNs, trained on segmented images, and local CNNs, trained on image patches. He et al. [50] presented a traffic sign recognition approach that integrates global and local features using histograms of oriented gradients (HOG), color histograms, and edge features. Suh et al. [51] employed fusion layers to concatenate global and local features for shipping label image classification, improving image quality verification. These studies demonstrated that fusing global and local features consistently improves classification performance compared to models relying on whole-image analysis alone.

In object detection, global features provide spatial awareness of objects within a scene, while local features capture subtle patterns, such as textures and edges, which are essential for accurate detection under occlusions. Li et al. [52] proposed Feature Fusion Single Shot Multibox Detector (FSSD), an enhanced version of SSD that incorporates a lightweight feature fusion module to better utilize multi-scale features. By concatenating features from different layers and applying down-sampling blocks, FSSD significantly improves detection accuracy with minimal speed loss, outperforming SSD and several state-of-the-art detectors on VOC and COCO benchmarks. Cong et al. [53] proposed an end-to-end co-salient object detection network that uses collaborative learning to enhance inter-image relationships. Their model includes a global correspondence module to extract interactive information across images and a local correspondence module to capture pairwise relationships. Later, Li et al. [54] developed an anchor-free object detector that uses a global-local feature extraction transformer (GLFT) to capture semantic information from both micro- and macro-level perspectives.

In semantic segmentation, the fusion of global and local features improves pixel-level predictions by combining overall scene context with localized information, especially in complex environments. For example, Zhang et al. [58] proposed ExFuse to address the semantic and resolution gap in fusing low-level and high-level features for semantic segmentation. By enriching low-level features with semantic context and high-level features with spatial detail, ExFuse significantly improved fusion effectiveness. Dai et al. [59] introduced Attentional Feature Fusion, a general framework that used multiscale channel attention to improve the fusion of features with inconsistent semantics and scales. By incorporating iterative attention mechanisms, their method addressed bottlenecks in conventional fusion strategies and achieved great performance on CIFAR-100 and ImageNet with fewer parameters, highlighting the effectiveness of attention-based fusion in deep networks. Yang et al. [60] introduced AFNet, which uses a multi-path encoder to extract diverse features, a multi-path attention fusion module, and a fine-grained attention fusion module to combine high-level abstract and low-level spatial features. Tian et al [61] extended this concept with two encoders to extract both global high-order interactive features and local low-order features. These encoders form the backbone of the global and local feature fusion network (GLFFNet), enabling effective segmentation of remote sensing images through a dual-encoder structure. Later, Zhou et al [62] proposed a local–global multi-scale fusion network (LGMFNet) for building segmentation in SAR images. LGMFNet includes a dual encoder-decoder structure, with a transformer-based auxiliary encoder complementing the CNN-based primary encoder. The global–local semantic aggregation module (GLSM) is also introduced to bridge the two encoders, enabling semantic guidance across multiple scales through a specialized fusion decoder.

Inspired by these advances, feature fusion techniques have gained traction in domain adaptation for Object Re-identification. Self-Similarity Grouping (SSG) [16] is the first approach to applied both global and local features for unsupervised domain adaptation (UDA) in Person ReID. However, SSG faces two challenges: first, using a single network for feature extraction often introduces noisy pseudo-labels, and second, it performs clustering independently on global and local features, potentially assigning multiple inconsistent pseudo-labels to the same sample. To address these limitations, $LF^2$ [17] was proposed to fuse global and local features into a unified representation, reducing noise and improving clustering consistency. Building on this idea, CORE-ReID [11] introduced an Ensemble Fusion module equipped with the ECAB, which effectively fuses global and local features.

## 3. Materials and Methods

Despite advancements in domain translation-based methods, these often suffer from a persistent domain gap between translated images and real target domain images, which can adversely impact performance. To address this issue, our approach employs a pseudo-labeling strategy, which enables data analysis at multiple levels of granularity. This method has demonstrated superior performance compared to domain translation-based techniques [11,15,44,45].

While existing pseudo-labeling frameworks such as Deep Mutual Learning (DML) [20], MMT [15], and MEB-Net [21] have proven effective, they suffer from limitations due to their heavy reliance on pseudo-labels generated by the teacher model. These pseudo-labels can be noisy or inaccurate, adversely affecting model training. To mitigate this issue, we utilize a teacher-student network paradigm, where the student network is trained on labeled source domain data, and the teacher network is iteratively refined using the Mean Teacher method. Furthermore, we incorporate the Ensemble Fusion++ module, which enhances feature extraction by adaptively refining both local and global representations, thereby it is expected to produce more stable and reliable pseudo-labels than existing approaches.

In CORE-ReID [11], the Efficient Channel Attention Block (ECAB) was primarily applied to local features, restricting the full potential of the Ensemble Fusion module. In CORE-ReID V2, we extend and enhance this module to ensure that both global and local features undergo comprehensive optimization. This enhancement results in a more balanced and discriminative feature representation, improving generalization across diverse ReID tasks. Additionally, the improved Ensemble Fusion++ is not only effective for Person ReID but also demonstrates strong domain adaptation capabilities in Vehicle ReID, further validating its versatility in Object ReID.

This chapter outlines the methodology and materials used in CORE-ReID V2 for unsupervised domain adaptation (UDA) in Object ReID. The proposed framework consists of two main stages: (1) pre-training on a labeled source domain and (2) fine-tuning on an unlabeled target domain.

*3.1. Overview*

3.1.1. CORE-ReID V1 and CORE-ReID V2

**CORE-ReID V1: A Baseline for Unsupervised Domain Adaptation in Person Re-identification**: CORE-ReID V1 was introduced as a framework to address Unsupervised Domain Adaptation (UDA) in Person Re-identification (ReID). It effectively tackled domain shifts between camera views by leveraging Camera-Aware Style Transfer for synthetic data generation, Random Grayscale Patch Replacement for data augmentation, and K-Means Clustering for pseudo-labeling. Additionally, the Ensemble Fusion module with Efficient Channel Attention Block (ECAB) played a crucial role in integrating local and global features, improving the model's performance in cross-domain scenarios.

Despite its success, CORE-ReID V1 had several limitations:

1. Limited Application Domain: The framework was specifically designed for Person ReID, restricting its applicability to other ReID tasks such as Vehicle ReID and Object ReID.
2. Synthetic Data Generation Challenge: The Camera-Aware Style Transfer method relied on predefined camera information, making it ineffective when the number of cameras was unspecified.
3. Inefficient Data Augmentation: The Random Grayscale Patch Replacement technique only operated locally, limiting its effectiveness in learning color-invariant features.
4. Clustering Limitations: The K-Means clustering used random centroid initialization, leading to poor centroid placement, slow convergence, high variance in clustering results, and imbalanced cluster sizes.
5. Feature Fusion Issue: The ECAB module enhanced only local features, neglecting improvements to global representations.
6. Restricted Backbone Support: The framework exclusively supported deep networks such as ResNet50, ResNet101, and ResNet152, making it computationally expensive and unsuitable for lightweight applications.

**CORE-ReID V2: Expanding Scope, Enhancing Performance:** To overcome these limitations, CORE-ReID V2 is proposed as a enhancement over CORE-ReID V1, expanding its capabilities to Vehicle ReID and Object ReID while introducing architectural and methodological improvements.

1. Expanded Application Scope: Unlike CORE-ReID V1, which was restricted to Person ReID, CORE-ReID V2 extends its applicability to Vehicle ReID and Object ReID, making it a versatile framework for various ReID tasks.
2. Advanced Synthetic Data Generation: CORE-ReID V2 incorporates both Camera-Aware Style Transfer and Domain-Aware Style Transfer, allowing effective synthetic data generation even when the number of cameras is unknown.

3. Improved Data Augmentation: A new grayscale patch replacement strategy considers both local grayscale transformation and global grayscale conversion, leading to better feature generalization across domains.
4. Enhanced Clustering with Greedy K-Means++: Instead of relying on random initialization, CORE-ReID V2 employs Greedy K-Means++, which selects optimized centroids to improve cluster spread; minimizes redundancy, requiring fewer iterations; enhances stability and consistency, reducing randomness; ensures better centroid distribution, leading to improved clustering performance.
5. Ensemble Fusion++ for Comprehensive Feature Enhancement: CORE-ReID V2 introduces Ensemble Fusion++, which integrates both ECAB and SECAB, ensuring that global features are enhanced alongside local features, leading to a more balanced and comprehensive feature representation.
6. Flexible Backbone Support: CORE-ReID V2 broadens its applicability by supporting lightweight networks such as ResNet18 and ResNet34, alongside ResNet50, ResNet101, and ResNet152. This allows deployment in computationally constrained environments, such as real-time and edge-based applications.

CORE-ReID V2 represents a substantial advancement over CORE-ReID V1 by expanding its scope beyond Person ReID, improving clustering stability, introducing adaptive feature enhancement mechanisms, and supporting lightweight architectures. Table 1 shows the summary of these improvements.

**Table 1.** Summary the main advancement of CORE-ReID V2 over CORE-ReID V1.

| Category | CORE-ReID V1 | | CORE-ReID V2 |
|---|---|---|---|
| | **Current Status** | **Drawbacks/ Issues** | |
| Applied Domain | Person ReID | Only support Person ReID. | Expansion from Person ReID to Vehicle ReID and further Object ReID. |
| Synthetic Data Generation | Camera-Aware Style Transfer | Do not work in case the number of cameras is not specified. | Camera-Aware Style Transfer and Domain-Aware Style Transfer (for the case the number of cameras is not specified). |
| Data Augmentation | Random gray scale patch replacement | Only replace random gray scale patch in the image locally. | Locally gray scale patch replacement and global gray scale conversion. |
| K-Means Clustering | Random initialization | Problems from random initialization (1) Poor centroid placement (2) Slow convergence (3) Stuck in local minima (4) High variance in results (5) Imbalanced cluster sizes | Greedy K-Means++ initialization helps: (1) Selects centroids with optimized spread (2) Minimizes redundancy, requiring fewer iterations (3) Improves initialization stability (4) Reduces randomness and provides consistent clusters (5) Ensures better centroid distribution |
| Ensemble Fusion | Ensemble Fusion with ECAB | Only the local features are enhanced in the Ensemble Fusion. | Ensemble Fusion++ (with ECAB and SECAB) helps enhance both local and global features. |
| Supported Backbones | ResNet50, 101, 152 | Do not support small backbones such as ResNet18, 34. | ResNet18, 34, 50, 101, 152 |

3.1.2. Problem definition and Methodology

**Problem definition**: We represent the customized labeled source domain data as $\mathbb{D}_S = \{(x_{S,i}, y_{S,i})|_{i=1}^{N_S}\}$, where $x_{S,i}$ and $y_{S,i}$ denote the $i^{th}$ source image and its corresponding ground truth identity label, respectively, and $N_S$ is the total number of source images. Similarly, the unlabeled target domain data is denoted as $\mathbb{D}_T = \{x_{T,i}|_{i=1}^{N_T}\}$, where $x_{T,i}$

indicates the $i^{th}$ target image, and $N_T$ is the number of target images. Identity labels are unavailable for the images in the target domain dataset, and it is important to note that the identities across the source and target domains do not overlap. The objective of Unsupervised Domain Adaptation (UDA) for Object ReID is to transfer knowledge from the source domain $S$ to the target domain $T$. To accomplish this, we propose the CORE-ReID V2 framework, designed to achieve effective knowledge transfer through a pseudo-label-based method.

**Methodology**: we adopt a pseudo-label-based approach by dividing the process into two stages: pre-training the model on the source domain using a fully supervised strategy, followed by fine-tuning it on the target domain through an unsupervised learning approach (Figure 1).

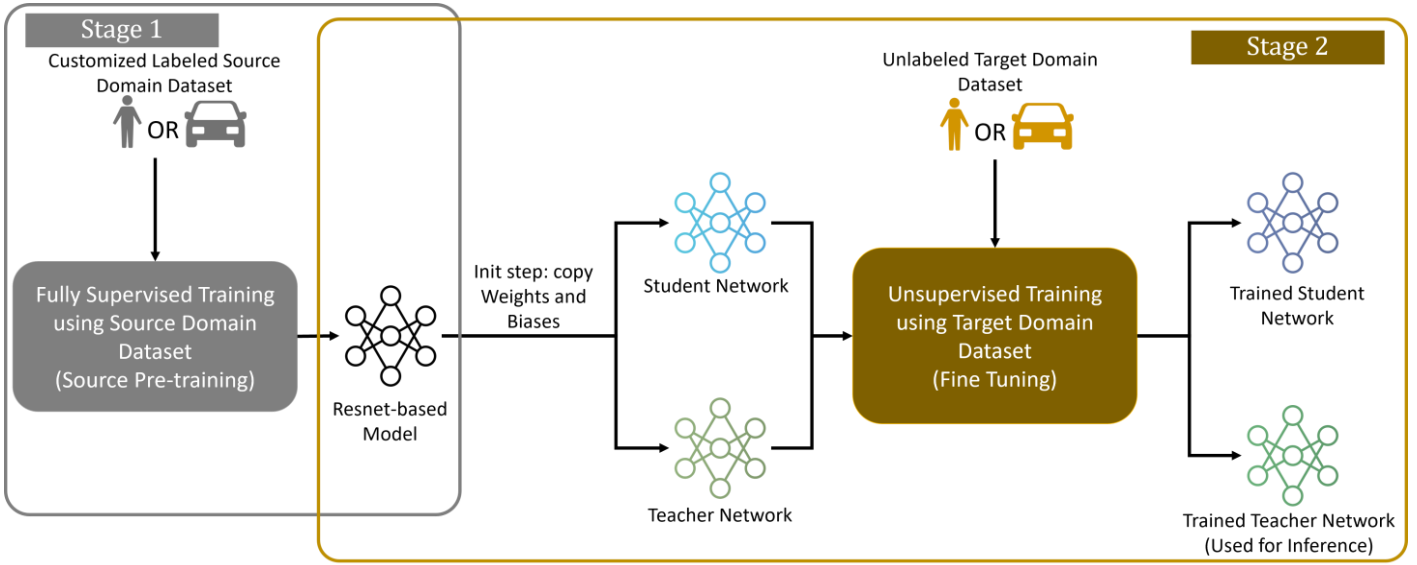

**Figure 1**. The overall method proposed in this study. First, the model is trained on a customized labeled source domain dataset, after which the parameters of the pre-trained model are transferred to both the student and teacher networks as an initialization step for the next stage. During fine-tuning, the student model is trained, and the teacher model is updated through the Mean Teacher method. To optimize computational efficiency, only the teacher model is employed for inference.

Depending on the specific task (Person ReID or Vehicle ReID), the appropriate dataset is utilized. Our method uses a pair of teacher–student networks. After training the model on a customized labeled source domain dataset, the parameters of the pre-trained model are copied to both the student and teacher networks as an initialization step for the fine-tuning stage. During fine-tuning, we first train the student model and then optimize the teacher model using the Mean Teacher method [22]. This is because averaging model weights across multiple training steps generally yields a more accurate model than relying solely on the final weights [86]. Following the Mean Teacher method, the teacher model uses Exponential Moving Average (EMA) weight parameters of the student model instead of directly sharing weights. This approach is expected to allow the teacher network to aggregate information after every step, rather than every epoch, improving consistency [22]. To minimize computational costs, only the teacher model is used during inference.

### 3.2. Source-domain pre-training

3.2.1. Image-to-Image translation

Inheriting from CORE-ReID, we employ CycleGAN to generate additional training samples by treating the stylistic variations across different cameras as distinct domains for Person ReID task. This involves training image-to-image translation models using CycleGAN for images captured from various camera views within the dataset. Our goal is to train on a source domain $S$ and evaluate the algorithm during the fine-tuning phase on a different target domain $T$. By incorporating test data into the training set, similar to DGNet++ [87], we can fully leverage the available data in $S$. In the Person ReID task, for a source domain dataset containing images from $C$ different cameras, we utilize $C(C-1)$ generative models to produce data in both $X \to Y$ and $Y \to X$ directions. The final training set is a combination of the original real images and the style-transferred images from both the training and test sets within the source domain dataset. These style-transferred images retain the labels of their corresponding real images.

In the case of Vehicle ReID, due to the simpler nature of vehicle features and the large number of cameras used (some datasets do not provide the number of cameras used), we adopt domain-aware transfer models instead of camera-aware models. As a result, only a single transfer model is needed to generate style-transferred images from the source domain to the target domain (Figure 2).

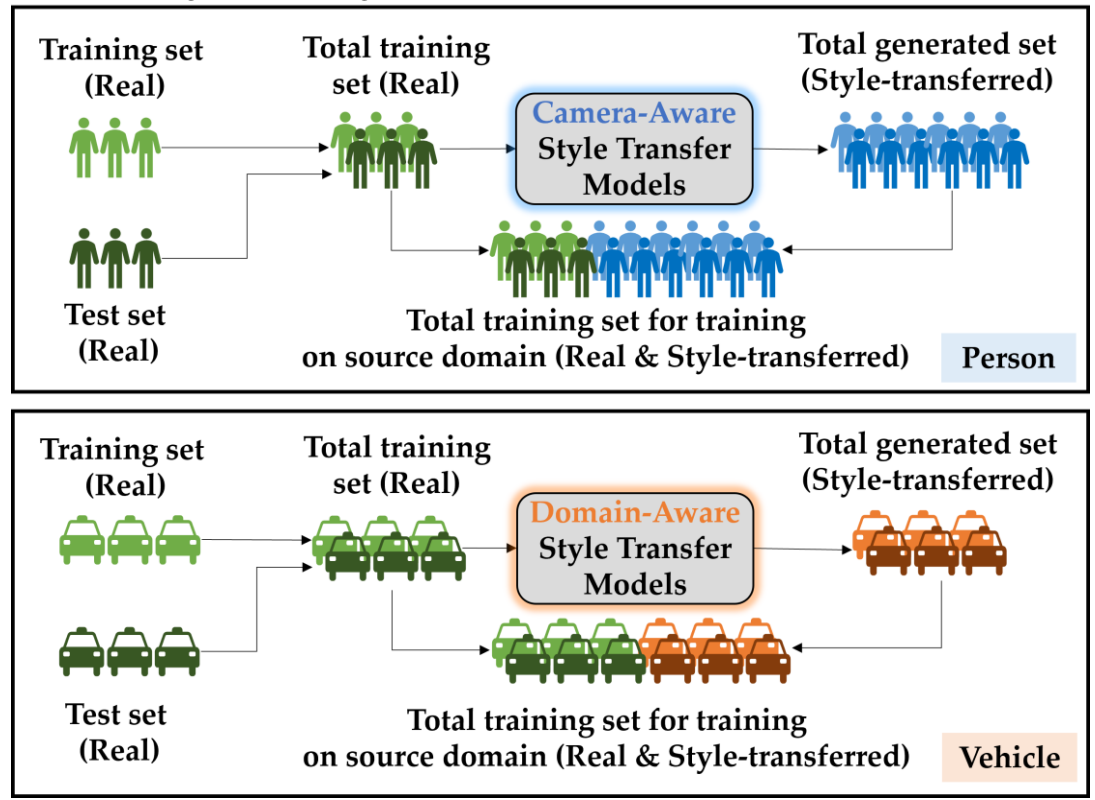

**Figure 2**. Our process for creating a complete training set for the source domain is as follows: in Person ReID task, we first combine the training set (represented by green icons) and the test set (represented by dark green icons) from the source dataset to create a comprehensive set of real images. This combined set is then used to train a camera-aware style transfer model, which generates style-transferred images (blue icons for the training set and dark blue icons for the test set) that reflect the stylistic characteristics of the target cameras. The final training set for the source domain is formed by merging the real images (green and dark green icons) with the style-transferred images (blue and dark blue icons). For Vehicle ReID, due to the simpler nature of vehicle features and the extensive number of cameras involved (with some datasets not specifying the number of cameras), we use domain-aware transfer models instead of camera-aware models. These models generate style-transferred images (orange icons for the training set and dark orange icons for the test set) that capture the target domain's style. The final source domain training set is then constructed by integrating the real images (green and dark green icons) with the style-transferred images (orange and dark orange icons).

Figure 3 illustrates two representative examples from both the training and test sets in the Market-1501 and CUHK03 datasets, where image styles have been modified according to camera views. This adjustment showcases our approach to data augmentation, where images are transformed to mimic the visual characteristics associated with each camera's unique viewpoint and color distribution. By aligning image styles with camera perspectives, this method effectively reduces the inconsistencies in appearance caused by differences in lighting, angles, and color shifts across camera views. This approach helps the model generalize better, thus mitigating overfitting in Convolutional Neural Networks (CNNs). Furthermore, incorporating camera-specific style information allows the model to learn more robust pedestrian features that are less sensitive to variations across different camera setups, leading to enhanced performance in ReID tasks.

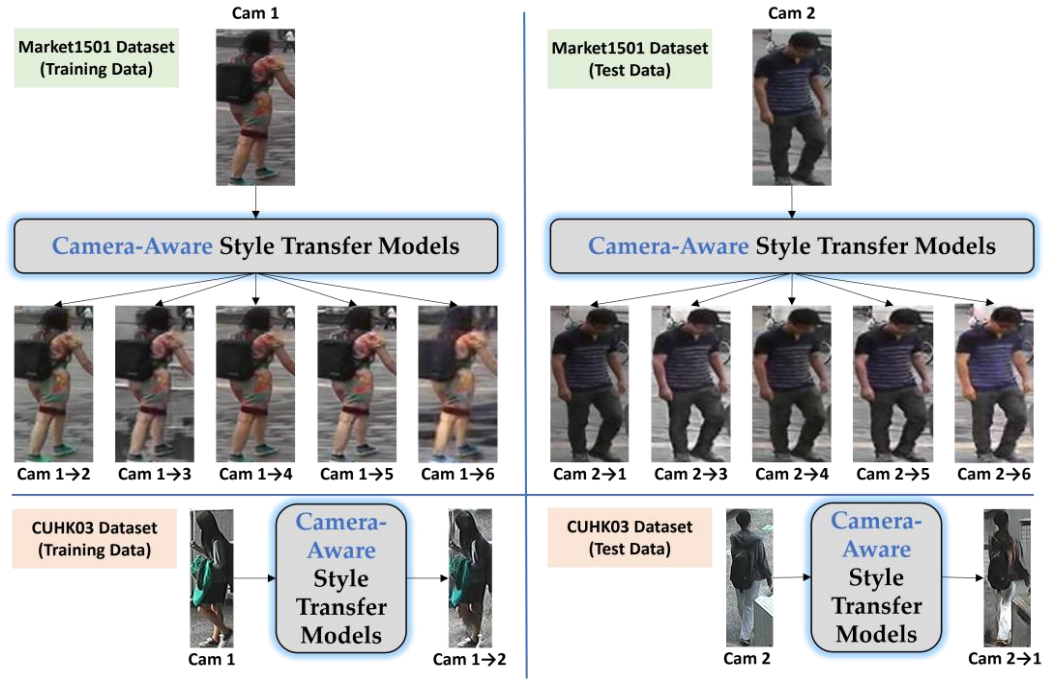

**Figure 3.** The camera-aware style-transferred samples from Market-1501 and CUHK03 datasets. Each original image, captured by a specific camera, has been transformed to match the styles of the other five cameras in Market-1501 and one camera in CUHK03, covering both training and test data. By applying the style transfer models, these transformations produce style-transferred images as outputs based on the real input images.

Given the simpler nature of vehicle features and the large number of cameras involved (with some datasets not specifying the number of cameras), we aim to utilize labels from the source domain along with target-domain-style-transferred images to create a shared-features domain dataset. This approach involves developing domain-aware style transfer models to bridge the feature gap between the two domains by transforming source domain images into target-domain-style outputs. Figure 4 shows two examples of input images from the VeRi-776 [1] and VehicleID [84] datasets, with the pixel differences between the input and output images highlighted in pink.

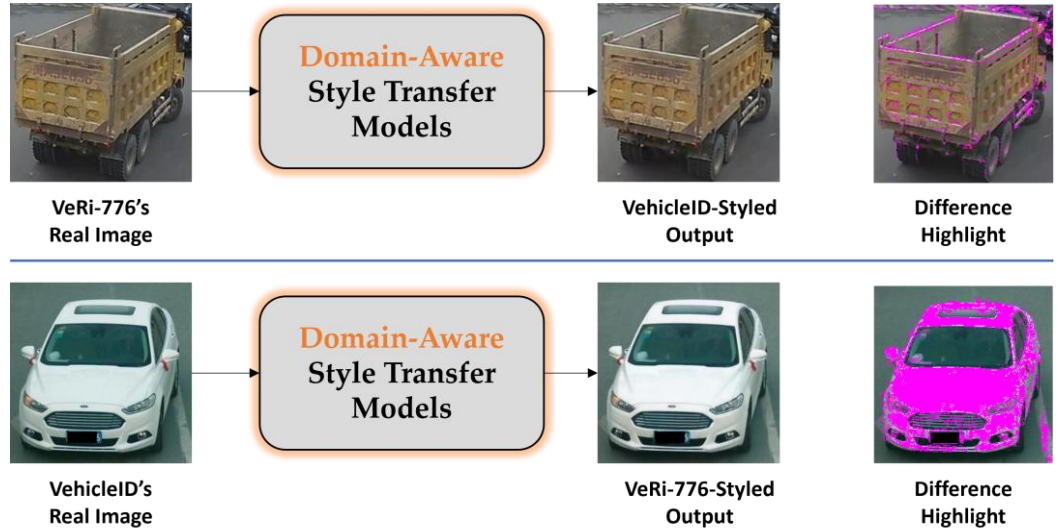

**Figure 4.** The domain-aware style-transferred samples from VeRi-776 [1] and VehicleID [84] datasets. Each original real image from the source domain dataset has been transformed to match the styles of the target domain dataset. By applying domain-aware style transfer models, the model trained in the pre−training stage will be able to capture the style and features of the target domain.

3.2.2. Fully supervised pre-training

Like many existing UDA approaches [17] that rely on a model pre-trained on a source dataset, we employ a ResNet-based model, pre-trained on ImageNet as the backbone network. In this setup, the original final fully connected (FC) layer is removed and replaced with two new layers. The first one is a batch normalization layer with either 2,048 or 512 features, depending on the specific ResNet architecture. The second layer is an FC layer with $M_S$ dimensions, where $M_S$ represents the number of identities (classes) in the source dataset $S$ (Figure 5).

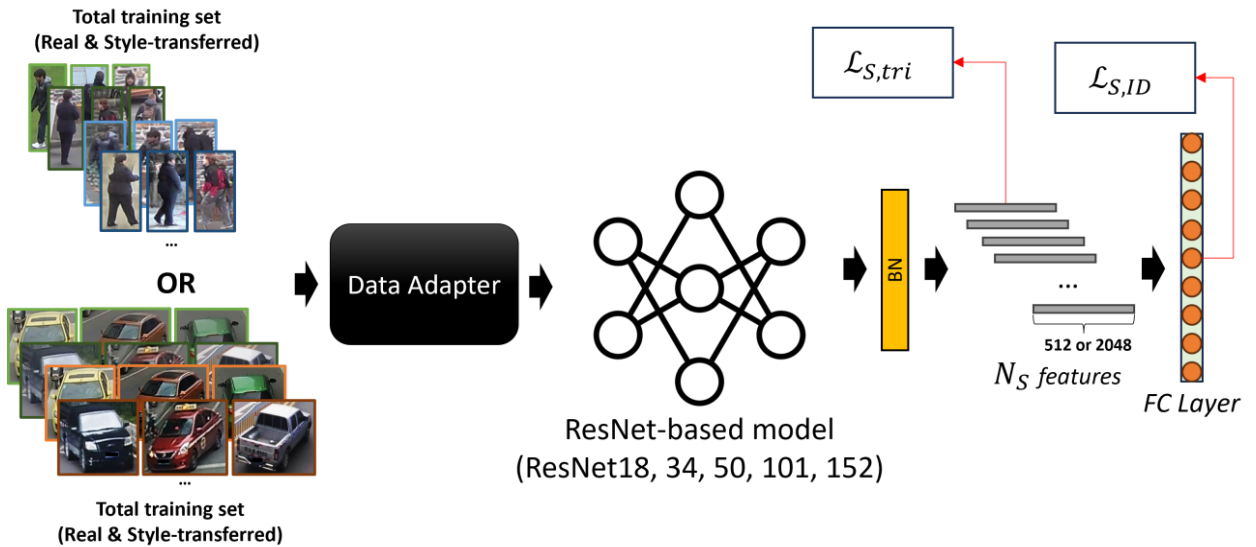

**Figure 5.** The comprehensive training process employed during the fully supervised pre-training stage. A ResNet-based model, adaptable to various backbone sizes (from ResNet18 and 34 to ResNet50, 101, and 150), serves as the backbone architecture within our training framework.

In our training process, we define the number of identities for the full training set in the source domain as:

$$M_{S,train} = M_{S,train}^{original} + M_{S,test}^{original}, \qquad (1)$$

where $M_{S,train}^{original}$ and $M_{S,test}^{original}$ represent the number of identities in the original training and test sets of $S$, respectively. For each labeled image $x_{S,i}$ and its ground truth identity $y_{S,i}$ in the source domain data $\mathbb{D}_S = \{(x_{S,i}, y_{S,i})|_{i=1}^{N_S}\}$ with $N_S$ representing the total number of images, we train the model using both identity classification (cross-entropy) loss $\mathcal{L}_{S,ID}$ and triplet loss $\mathcal{L}_{S,triplet}$. The identity classification loss is applied to the final fully connected (FC) layer, handling the task as a classification problem, while the triplet loss, applied after batch normalization, is used for feature verification. The loss functions are defined as follows:

$$\mathcal{L}_{S,ID} = \frac{1}{N_S} \sum_{i=1}^{N_S} \mathcal{L}_{ce}(C_S(f(x_{S,i}), y_{S,i}), \quad (2)$$

$$\mathcal{L}_{S,triplet} = \frac{1}{N_S} \sum_{i=1}^{N_S} \max(0, ||f(x_{s,i}) - f(x_{S,i}^+)||_2 - ||f(x_{s,i}) - f(x_{S,i}^-)||_2 + m), \quad (3)$$

where $f(x_{S,i})$ is the feature embedding vector of the source image $x_{S,i}$ extracted from the network with the backbone in Figure 5, $\mathcal{L}_{ce}$ is the cross-entropy loss, $C_S$ is a learnable classifier in the source domain: $f(x_{S,i}) \rightarrow \{1,2,...,M_S\}$. $||\cdot||_2$ denotes the $L_2$-norm distance, $x_{S,i}^+$ and $x_{S,i}^-$ are the hardest positive and hardest negative feature indices in each mini-batch for the sample $x_{S,i}$, and $m$ represents the triplet distance margin. Using a balance parameter $\kappa$, the total loss for source-domain pre-training is:

$$\mathcal{L}_{S,total} = \mathcal{L}_{S,ReID} = \mathcal{L}_{S,ID} + \kappa \mathcal{L}_{S,triplet}. \quad (4)$$

The model is expected to achieve strong performance on fully labeled source-domain data, but its performance significantly drops when applied directly to the unlabeled target domain. Before feeding images into the network, we utilize the "Data Adapter" component (Figure 6) to preprocess them by resizing to a specific size depending on the type of object. We then apply several data augmentation techniques, including edge padding, random cropping, and random horizontal flipping. To address color deviation, we incorporate random color dropout through global and local grayscale transformations [88], preserving key information while minimizing overfitting and enhancing the model's generalization. These approaches specifically balance the model's weighting of color features and color-independent features, resulting in improved feature robustness in the neural network.

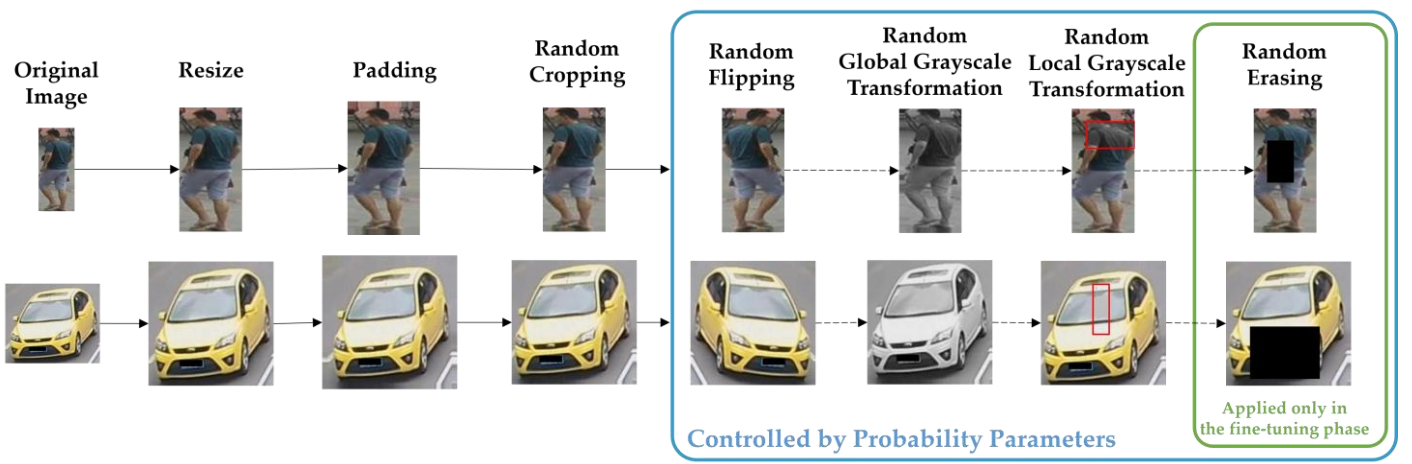

**Figure 6.** Data adapter component. The transformations of random flipping, random global grayscale, random local grayscale and random erasing will be controlled by probability parameters. In addition, random erasing is only applied in the fine-tuning stage.

We employ the global grayscale transformation to a training batch with a set probability $p_{global}$, then feed it into the model for training. This process is defined as: $I^* = RGBToGrayscale(I)$, where $RGBToGrayscale()$ represents the grayscale conversion function using the NTSC formula ($0.299 \times Red + 0.587 \times Green + 0.114 \times Blue$), $I$ denotes the input image and $I^*$ is the randomly grayscale image. This function operates by performing pixel-wise accumulation on the red, green, and blue channels of the original RGB image, resulting in a grayscale output. Importantly, the labels remain consistent between the converted grayscale image and the original. The procedure for Local Grayscale Transformation is outlined in Algorithm 1.

---
**Algorithm 1:** Global Grayscale Transformation
---
**Input:** Input image $I$;
    Grayscale transformation probability $p_{global}$.
**Output:** Randomly grayscale image $I^*$.
**Initialization:** $p_t \coloneqq Rand\,(0,1)$.
  1: **if** $p_t \geq p_{global}$ **then**
  2:    $I^* \coloneqq I$.
  3: **else**
  4:    $I^* \coloneqq RGBToGrayscale(I)$.
  5: **return** $I^*$.
  6: **end**

---

To enhance model adaptability to significant biases from localized color dropout, we apply a local grayscale transformation to each visible image $I$ in the training batch using the following equation:

$$I^*_{position} = RGBToGrayscale\left(I_{position}\right) = RGBToGrayscale(RandPosition(I)), \quad (5)$$

where $RandPosition()$ generates a random rectangular region within the image $I$, The transformed sample is represented by $I^*$. During model training, local grayscale transformation is applied randomly to images in each batch with a probability $p_{local}$. This involves selecting a random rectangular region within the image and replacing it with the grayscale pixels of that same region. Consequently, images with mixed grayscale levels are generated, aiding the model in learning with color-variant features without altering object structure. The process includes several parameters: $s_{min}$ and $s_{max}$ define the minimum and maximum size ratios of the rectangle relative to the full image area; the rectangle's area $S_t$ is computed by sampling from $S_t \leftarrow Rand(s_{min}, s_{max}) \times S$, where $S$ is the input image area; $r_t$ is a coefficient that sets the rectangle's shape ratio within the interval ($r_{local}$, $1/r_{local}$); coordinates $x_t$ and $y_t$ for the rectangle's top-left corner are generated randomly. If the rectangle exceeds image boundaries, new coordinates and dimensions are selected. This approach produces images with grayscale sections that vary in intensity without impacting the core structure, allowing the model to learn features invariant to color variations. The full procedure for local grayscale transformation is detailed in Algorithm 2.

---
**Algorithm 2:** Local Grayscale Transformation
---
**Input:** Input image $I$;
    Grayscale transformation probability $p_{local}$;
    Area ratio range (low to high) $s_{min}$ and $s_{max}$;
    Aspect ratio $r_{local}$.
**Output:** Randomly transformed image $I^*$.
**Initialization**: $p_t \coloneqq Rand(0,1)$;
        $W \coloneqq I.size[0]$, $H \coloneqq I.size[1]$;
        $S \coloneqq W * H$.
  1: **if** $p_t \geq p_{local}$ **then**
  2:    $I^* \coloneqq I$;
        **return** $I^*$.

```
 3:  else
 4:    while True do
 5:       S_t := Rand(s_min, s_max) × S;
 6:       r_t := Rand(r_local, 1/r_local);
 7:       W_t := √(S_t/r_t) , H_t := √(S_t × r_t);
 8:       x_t := Rand(0, W), y_t := Rand(0, H);
 9:       if x_t + W_t ≤ W and y_t + H_t ≤ H then
10:          I_position := (x_t, y_t, x_t + W_t, y_t + H_t);
11:          I_position := RGBToGrayscale (I_position);
12:          I* := I;
13:          return I*.
14:       end
15:    end
16:  end
```

3.2.3. Implementation details

To perform camera-aware image-to-image translation for generating synthetic data, we train 30 generative models for the Market-1501 dataset and 2 models for the CUHK03 dataset. These numbers are derived from the formulas $6 \times (6 - 1) = 30$ and $2 \times (2 - 1) = 2$, corresponding to the number of camera pairs in each dataset. For domain-aware image-to-image translation, we train 2 generative models (VeRi-776 to VehicleID and reverse). During training, all input images are first resized to 286×286 pixels, followed by cropping them to 256×256 pixels. We use the Adam optimizer for training all models from scratch, with a batch size of 8. The learning rate is initialized at 0.0002 for the Generator and 0.0001 for the Discriminator. For the first 30 epochs, these rates are kept constant and then linearly decayed to near zero over the subsequent 20 epochs according to a lambda learning rate schedule.

For pre-training, we adopt ResNet101 as the backbone (with support for other ResNet architectures as well). The initial learning rate is set to 0.00035, then reduced by a factor of 0.1 at the 40[th] and 70[th] epochs, totaling 350 training epochs with a 10-epoch warmup period. Each training batch consists of 32 identities, with 4 images per identity, resulting in a final batch size of 128. The balance parameter κ for computing the total loss is set to 1. Regarding preprocessing, each image is resized to 256×128 pixels for Person ReID task and 256×256 pixels for Vehicle ReID task. The resized images are padded with 10 pixels using edge padding, followed by random cropping back to their original resized dimensions. Additional augmentation techniques included random horizontal flipping, global grayscale transformation $p_{global}$, and local grayscale transformation $p_{local}$, applied with probabilities of 0.5, 0.05, and 0.4, respectively. Images are then converted to 32-bit floating-point pixel values normalized to the $[0,1]$ range. The RGB channels are further normalized by subtracting mean values of $[0.485, 0.456, 0.406]$ and dividing by standard deviations of $[0.229, 0.224, 0.225]$.

3.3. Target-domain fine-tuning

3.3.1. Overall algorithm

In this phase, we use the pre-trained model to perform comprehensive optimization. We present our CORE-ReID V2 framework (Figure 7) along with Efficient Channel Attention Block (ECAB) and Simplified Efficient Channel Attention Block (SECAB) in Ensemble Fusion++.

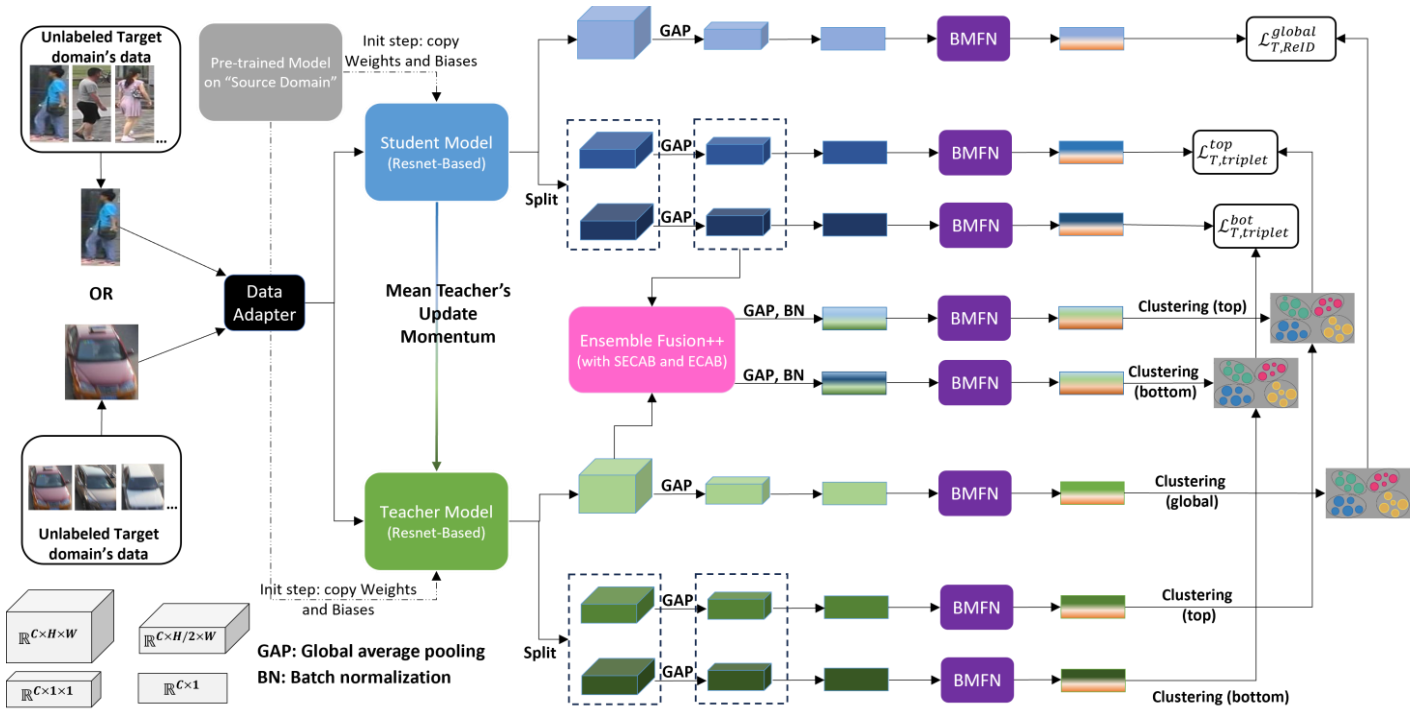

**Figure 7.** The comprehensive overview of our CORE-ReID V2 framework. The data adapter will pre-process the data depending on the type of object. We integrate local and global features using the enhanced Ensemble Fusion++ component. Specifically, the Efficient Channel Attention Block (ECAB) and Simplified Efficient Channel Attention Block (SECAB) are utilized to boost local and global feature extraction, respectively. By using Bi-directional Mean Feature Normalization (BMFN), the framework effectively merges features from the original image $x_{T,i}$ and its horizontally flipped counterpart $x'_{T,i}$, generating a fused feature $\varphi_l, l \in \{top, bottom\}$. The student network is trained in a supervised manner using pseudo-labels, while the teacher network is updated through a Mean Teacher approach, which computes the temporal average of the student network's weights. Especially, the flipped image features are processed identically to the original image's features until they reach the BMFN stage, ensuring consistent feature fusion.

Building upon the strategies utilized in SSG [16], LF[2] [17], and CORE-ReID [11], our objective is to enable the model to dynamically integrate both global and local features. This approach allows for feature representations that encompass comprehensive global and detailed local information. To further enhance these feature representations, we incorporate ECAB and SECAB modules during the fusion process. By organizing multiple clusters based on global and fused features, we aim to generate more reliable pseudo-labels, thereby reducing the risk of ambiguous learning.

To refine these pseudo-labels, we implement a teacher-student network pair grounded in the mean-teacher framework. We feed the same unlabeled image from the target domain into both the teacher and student networks. During iteration $i$, the student network's parameters, $\rho_\varsigma$, are updated using Mean Teacher momentum, adjusting them through backpropagation within the target domain training. In parallel, the teacher network's parameters, $\rho_\tau$, are derived as a moving average of the student network's parameters $\rho_\varsigma$. This is controlled by the temporal momentum coefficient $\eta$, which is restricted to the range $[0,1)$. The update rule is defined as:

$$\rho_{\tau,i} = \eta \rho_{\tau,i-1} + (1-\eta)\rho_\varsigma, \tag{6}$$

We employ the K-means clustering algorithm to assign pseudo-labels to the data, using the Euclidean distance as the similarity metric. As a result, each sample $x_{T,i}$ is assigned three pseudo-labels (global, top, and bottom). The target domain dataset is defined as: $\mathbb{D}_T = \{(x_{T,i}, \hat{y}_{T,i,j})\big|_{i=1}^{N_T}\}$, where $j \in \{global, top, bottom\}$ and $N_T$ represents the total number of images in the target dataset $T$. The pseudo-label $\hat{y}_{T,i,j} \in$ $\{1,2\ldots,M_{T,j}\}$ indicates that $\hat{y}_{T,i,j}$ is derived from the clustering results $\hat{Y}_j = \{\hat{y}_{T,i,j} | i = 1,2,\ldots,N_T\}$. These are obtained using the combined feature with its flipped counterpart $x'_{T,i}$ generated by BMFN, denoted as $\varphi_l, l \in \{top, bottom\}$. Here, $M_{T,j}$ stands for the number of distinct identities (classes) in the clustering outcome $\hat{Y}_j$.

Before computing the loss function, we use BMFN to extract optimized features from networks $f_j^\varsigma$, $f_j^\tau$, $j \in \{global, top, bottom\}$ and $\varphi_l, l \in \{top, bottom\}$ from the Ensemble Fusion++. Given an image $x_{T,i}$ in the target dataset, along with its flipped version $x'_{T,i}$, we extract the feature maps $F_j^m$ and the flipped feature maps $F'^m_j$ for $j \in \{global, top\ bottom\}$ and $m \in \{\varsigma, \tau\}$. The BMFN output is computed as follows:

$$f_j^m = \text{BMFN}(F_j^m, F'^m_j) = \frac{\frac{F_j^m + F'^m_j}{2}}{\|\frac{F_j^m + F'^m_j}{2}\|_2}. \tag{7}$$

$$\varphi_l^m = \text{BMFN}(\theta_l^m, \theta'^m_l) = \frac{\frac{\theta_l^m + \theta'^m_l}{2}}{\|\frac{\theta_l^m + \theta'^m_l}{2}\|_2}, \tag{8}$$

After obtaining multiple pseudo-labels, we generate three new target-domain datasets to train the student network. The pseudo-labels derived from the local fusion features, denoted as $\varphi_l, l \in \{top, bottom\}$, are used to calculate the softmax triplet loss for the corresponding local features $f_l^\varsigma$ from the student network:

$$\mathcal{L}_{T,triplet}^l = \frac{1}{N_T} \sum_{i=1}^{N_T} \log\left(\frac{e^{\|f_l^\varsigma(x_{T,i}|\rho_\varsigma) - f_l^\varsigma(x_{T,i}^-|\rho_\varsigma)\|_2}}{e^{\|f_l^\varsigma(x_{T,i}|\rho_\varsigma) - f_l^\varsigma(x_{T,i}^-|\rho_\varsigma)\|_2} + e^{\|f_l^\varsigma(x_{T,i}|\rho_\varsigma) - f_l^\varsigma(x_{T,i}^+|\rho_\varsigma)\|_2}}\right), \tag{9}$$

where $\rho_\tau$ and $\rho_\varsigma$ are the parameters of the teacher and student networks, respectively. The optimized local feature from the student network is denoted as $f_l^\varsigma$, with $l \in \{top, bottom\}$. Here, $x_{T,i}^+$ and $x_{T,i}^-$ represent the hardest positive and negative samples relative to the anchor image $x_{T,i}$ in the target domain.

In a similar fashion to supervised learning, we utilize the cluster results $\hat{Y}_{global}$ of the globally clustered feature $f_{global}^\varsigma$ as pseudo-labels to compute the classification loss $\mathcal{L}_{T,ReID}^{global}$ and the global triplet loss $\mathcal{L}_{T,triplet}^{global}$. These losses are defined as follows:

$$\mathcal{L}_{T,ID}^{global} = \frac{1}{N_T} \sum_{i=1}^{N_T} \mathcal{L}_{ce}(C_T(f_{global}^\varsigma(x_{T,i}), \hat{y}_{T,i,global}), \tag{10}$$

$$\mathcal{L}_{T,triplet}^{global} = \frac{1}{N_T} \sum_{i=1}^{N_T} \max(0, \|f_{global}^\varsigma(x_{T,i}) - f_{global}^\varsigma(x_{T,i}^+)\|_2 - \|f_{global}^\varsigma(x_{T,i}) - f_{global}^\varsigma(x_{T,i}^-)\|_2 + m), \tag{11}$$

where $C_T$ represents the fully connected classification layer of the student network, mapping $f_{global}^\varsigma(x_{T,i})$ to the set $\{1,2,\ldots,M_{T,global}\}$. The notation $\|\cdot\|_2$ indicates the $L_2$-norm distance.

The total loss is computed by combining the different losses with weighting parameters $\alpha, \beta, \gamma$:

$$\begin{aligned}\mathcal{L}_{T,total} &= \mathcal{L}_{T,ReID}^{global} + \gamma \mathcal{L}_{T,triplet}^{top} + \delta \mathcal{L}_{T,triplet}^{bottom} \\ &= \alpha \mathcal{L}_{T,ID}^{global} + \beta \mathcal{L}_{T,triplet}^{global} + \gamma \mathcal{L}_{T,triplet}^{top} + \delta \mathcal{L}_{T,triplet}^{bottom}.\end{aligned} \tag{12}$$

During the inference phase, the Ensemble Fusion++ process is bypassed, using only the optimized teacher network to reduce computational overhead. Specifically, the global feature map from the teacher network is split into two segments, referred to as top and bottom features (which also acts similarly in the student network). These segments undergo global average pooling, after which the two local features and the global feature are concatenated. Finally, $L_2$ normalization and the BMFN method are applied to obtain the optimal feature representation for inference.

3.3.2. Ensemble Fusion++ component

To extract the fusion features, we horizontally divide the final global feature map of the student network into two segments (top and bottom), resulting in $\varsigma_{top}$ and $\varsigma_{bottom}$ after applying global average pooling. Unlike the Ensemble Fusion component in CORE-ReID [Nguyen, 2024 #29], the final global feature map $\tau_{global}$ from the teacher network is further enhanced using the proposed SECAB module. These features $\varsigma_{top}$ and $\varsigma_{bottom}$ from the student network, along with $\tau_{global}$ from the teacher network are then utilized for adaptive feature fusion through the Ensemble Fusion++ module, which includes learnable parameters.

The inputs $\varsigma_{top}$ and $\varsigma_{bottom}$ are processed by ECAB, while $\tau_{global}$ is processed by SECAB for adaptive fusion. The enhanced attention maps ($\psi_{top}$ and $\psi_{bottom}$) generated by ECAB are combined with the output $\tau'_{global}$ through element-wise multiplication, resulting in the ensemble fusion feature maps: $\tau'^{top}_{global}$ and $\tau'^{bottom}_{global}$. These maps undergo Global Average Pooling (GAP) and batch normalization, yielding the fusion features $\theta_{top}$ and $\theta_{bottom}$. These features are then fed into the BMFN for predicting pseudo-labels using clustering algorithms in subsequent steps.

The process within Ensemble Fusion++ (Figure 8) can be summarized as follows:

$$\tau'^{top}_{global} = \psi_{top} \otimes \tau'_{global} = ECAB(\varsigma_{top}) \otimes [\tau_{global} \otimes SECAB(\tau_{global})], \quad (13)$$

$$\tau'^{bot}_{global} = \psi_{bottom} \otimes \tau'_{global} = ECAB(\varsigma_{bottom}) \otimes [\tau_{global} \otimes SECAB(\tau_{global})], \quad (14)$$

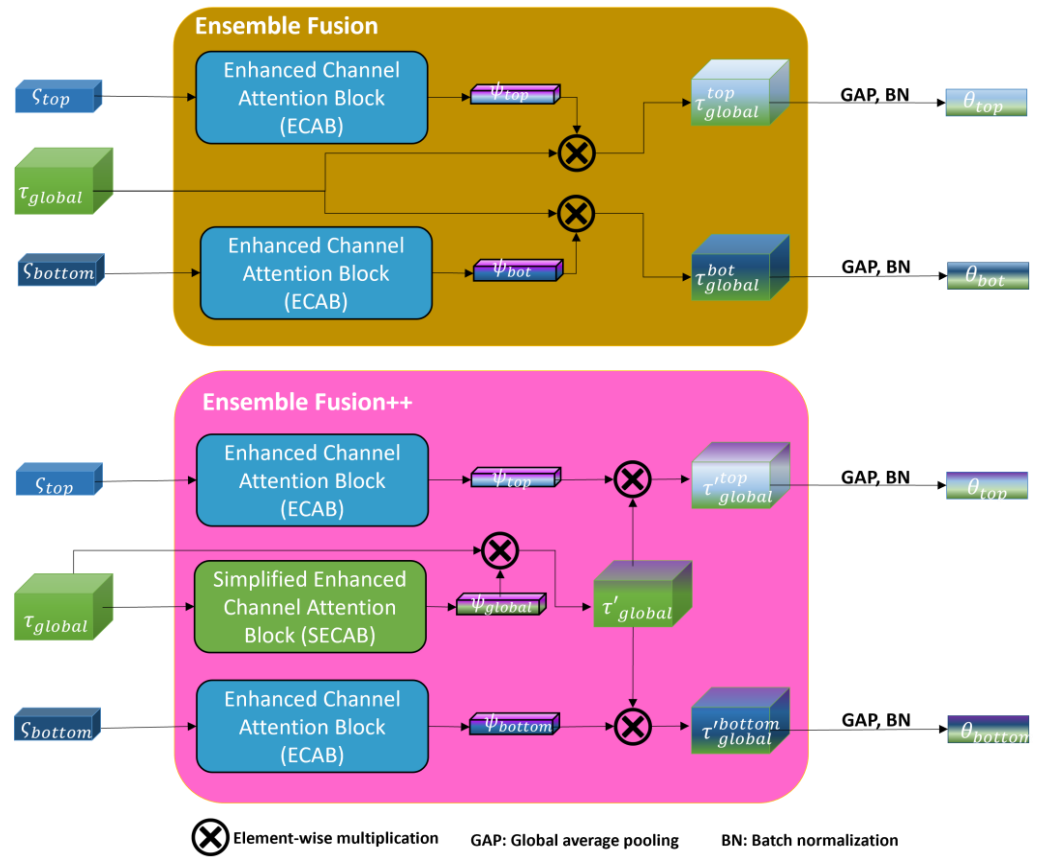

**Figure 8.** The comparison between Ensemble Fusion in CORE-ReID [11] and proposed Ensemble Fusion++ component. $\varsigma_{top}$ and $\varsigma_{bottom}$ features are passed through the ECAB, $\tau_{global}$ feature is passed via the SECAB to produce the channel attention maps by exploiting the inter-channel relationship of features which helps to enhance the features.

3.3.3. SECAB

The importance of attention has been extensively explored in previous literature [89] [90] [91]. Attention not only guides where to focus but also enhances the representation of relevant features. Inspired by the ECAB [11], we introduce a new component named SECAB, a straightforward yet impactful attention module for feed-forward convolutional neural networks to enhance the global feature. ECAB is designed to refine channel-wise feature representations by using both max-pooling and average-pooling operations, followed by a Shared Multilayer Perceptron (SMP) with ReLU activations. This SMP enhances non-linearity, and models complex inter-channel dependencies more effectively. After generating attention maps from both pooled features, ECAB applies a sigmoid activation and multiplies the result with the sum of the pooled inputs, producing a strongly refined output. In contrast, the SECAB (Simplified ECAB) is a lightweight, GPU-friendly variant specifically tailored for global feature refinement. While it retains the same attention generation pathway (pooling → SMP → sigmoid), it omits the reweighting step and directly outputs the attention map. This reduces computational complexity while still maintaining meaningful channel attention on global features. Figure 9 shows the design of SECAB, while Table 2 describes the comparison of ECAB and SECAB.

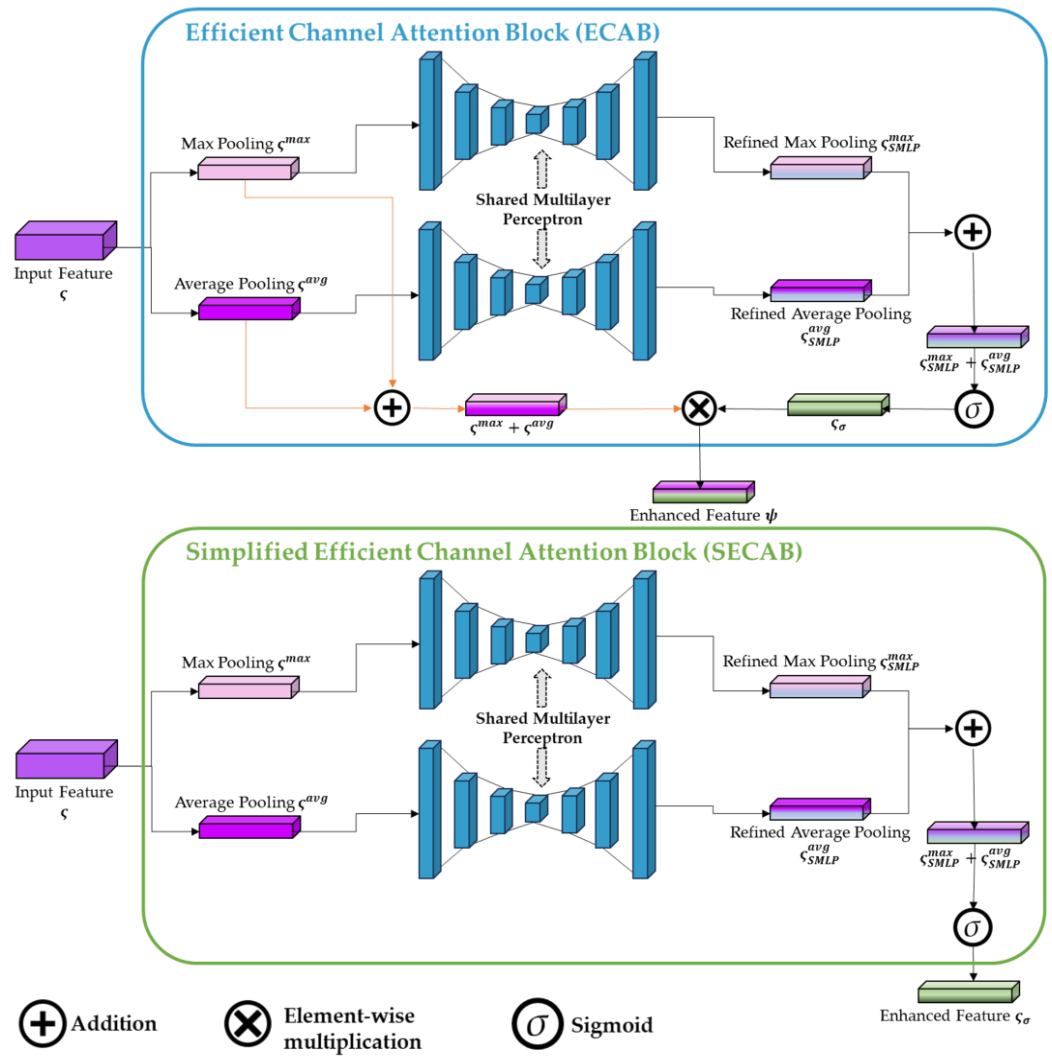

**Figure 9.** ECAB and SECAB designs. The structure of our SECAB is similar to ECAB [11] but simpler, the module only takes the Shared Multilayer Perceptron into account. It has odd $h$ hidden

layers, where the first $\frac{h-1}{2}$ layers are reduced in size with the reduction rate $r$, and the last $\frac{h-1}{2}$ layers will be expanded with the same rate $r$.

**Table 2.** Comparison of ECAB and SECAB.

| Aspect | ECAB | SECAB |
| --- | --- | --- |
| Target Use | Local feature vectors | Global feature map |
| Pooling | Adaptive Max + Avg Pooling | Adaptive Max + Avg Pooling |
| Attention Core | Shared Multilayer Perceptron | Same Shared Multilayer Perceptron |
| Output Processing | Attention map × (max + avg feature) | Attention map only |
| Residual Information Fusion (Later) | With refined global features | With original global features |
| Computational Cost | Higher (due to residual and additional element-wise operations) | Lower (no fusion step, lightweight on GPU) |
| Deployment Stage | Local-level features refinement | Global-level features refinement |
| Used in Ensemble Fusion | Yes | No |
| Used in Ensemble Fusion++ | Yes | Yes |

Given an intermediate input feature map $\varsigma \in \mathbb{R}^{C \times W \times H}$, where $C, W, H$ denote the number channel, width, and height respectively. After performing max-pooling and average-pooling, then fit the outputs $\varsigma^{max}$, $\varsigma^{avg}$ into a Shared Multilayer Perceptron (SMP), we can obtain refined feature as $\varsigma^{max}_{SMLP}$ and $\varsigma^{avg}_{SMLP}$. The SMP has multiple hidden layers with reduction rate $r$ and the same expansion rate, activation function ReLU. Sigmoid activation squashes the sum to the range between 0 and 1, producing a channel-wise attention mask. The enhanced attention map $\varsigma_\sigma \in \mathbb{R}^{C \times 1 \times 1}$ is calculated as:

$$\varsigma_\sigma = \sigma(\varsigma^{max}_{SMLP} + \varsigma^{avg}_{SMLP}), \qquad (15)$$

3.3.4. Greedy K-means++

In the K-means clustering problem, we are given a set of points $N_T \subseteq \mathbb{R}^{dim}$ in a $di$-dimensional space, a specified number of clusters $M_{T,j}$. The objective is to identify a set of $M_{T,j}$ centroids $C = \{c_1, c_2, \ldots, c_{M_{T,j}}\} \subseteq \mathbb{R}^{dim}$ that minimizes the total sum of squared distances from each point in $N_T$ to its nearest centroid. Specifically, if we define the cost of a point $x$ with respect to a set of centers $C$ as $\mathcal{D}(x, C) \coloneqq min_{c \in C} \|x - c\|^2$, the goal is to find $C$ such that $|C| = M_{T,j}$ and the cost $\mathcal{D}(N_T, C) \coloneqq \sum_{x \in N_T} \mathcal{D}(x, C)$ is minimized. In practice, a simple way to initialize centroids is to select a random subset of $N_T$ with the size $M_{T,j}$. However, this random approach does not guarantee any approximation bounds and can perform poorly in certain cases, such as when $M_{T,j}$ well-separated clusters exist along a single line. Arthur and Vassilvitskii [19] propose a probabilistic seeding method known as K-means++, which improves centroid initialization by favoring points that are far from already-selected centroids, while still maintaining a degree of randomness. Empirical results demonstrate that K-means++ consistently outperforms random seeding on real-world datasets [19].

The Greedy K-means++ algorithm [92] refines this process further by eliminating randomness and selecting centroids deterministically to explicitly maximize the spatial spread. The algorithm operates as follows. At each step, it samples $\ell$ candidate $c^1_{i+1}, c^2_{i+1}, \ldots, c^\ell_{i+1}$ from a distribution constructed based on the current centroid configuration. Then, for each candidate $c^j_{i+1}$, the algorithm calculates the new cost $\mathcal{D}(X, C \cup \{c^j_{i+1}\})$ that would result from adding this candidate to the set of centroids. Next, the candidate center that minimizes this cost is selected as the next centroid. In our implementation, $\ell$ is typically set to $2 + \log(M_{T,j})$. By systematically evaluating multiple candidates at each step, Greedy K-means++ ensures better centroid initialization and improved clustering outcomes (Algorithm 3).

---

**Algorithm 3:** Greedy K-means++ seeding

**Input:** The number of images in target dataset $N_T$;
   The number of the clusters $M_{T,j}$;
   The number of candidate centers $\ell$.
**Output:** The set of centers $C$.
**Initialization**: Uniformly independently sample $c_1^1, c_1^2, \ldots, c_1^\ell \in N_T$.
1: Let $c_1 = argmin_{c \in \{c_1^1, c_1^2, \ldots, c_1^\ell\}} \mathcal{D}(X, c)$ and set $C_1 = \{c_1\}$.
2: **for** $i \leftarrow 1, 2, \ldots, M_{T,j} - 1$ **do**
3:   Sample $c_{i+1}^1, c_{i+1}^2, \ldots, c_{i+1}^\ell \in N_T$ independently;
4:   Sample $x$ with probability $\frac{\mathcal{D}(x, C_i)}{\mathcal{D}(X, C_i)}$ ;
5:   Let $c_{i+1} = argmin_{c \in \{c_i^1, c_i^2, \ldots, c_i^\ell\}} \mathcal{D}(X, C_i \cup \{c\})$;
6:   Set $C_{i+1} = C_i \cup \{c_{i+1}\}$.
7: **return** $C := C_{M_{T,j}}$

---

3.3.4. Detailed implementation

The training process lasts 80 epochs, with each epoch consisting of 400 iterations. A fixed learning rate of 0.00035 for Person ReID task (0.00007 for Vehicle ReID task) is maintained throughout, and the Adam optimizer is employed with a weight decay of 0.0005 to ensure stable convergence. Clustering operations utilize the K-means algorithm with Greedy K-means++ initialization, where the maximum number of iterations is capped at 100, striking a balance between computational efficiency and solution accuracy. The mini-batch size is set to 512, allowing for efficient centroid updates without processing the entire dataset. An early stopping criterion is applied, terminating clustering if no improvement in inertia is observed over 50 consecutive mini-batches. To address the issue of empty clusters, a reassignment ratio of 0.05 is used, ensuring toughness in dynamic data distributions. For centroid initialization, 1,500 data points are used for global features, while 900 data points are allocated for both top and bottom local features.

In the temporal ensemble regularization process, we follow the common practice from the original Mean Teacher paper [22] and set the momentum parameter ($\eta$) to 0.999, which has been shown to work well in practice across various tasks. To balance the contributions of the various components in the loss function, we assign weights as follows: $\alpha = 1, \lambda = 1, \gamma = 0.5$, and $\delta = 0.5$. For the Ensemble Fusion++ module, a reduction ratio and expansion rate ($r$) of 4 are utilized, along with 5 hidden layers ($h$) for both ECAB and SECAB components.

The data adapter resizes input images to $128 \times 256$ for the Person ReID task and $256 \times 256$ for the Vehicle ReID task. Edge padding of 10 pixels is applied before randomly cropping the images to their respective dimensions ($128 \times 256$ or $256 \times 256$). Data augmentation strategies include random horizontal flipping, global grayscale transformation, local grayscale transformation, and random erasing, applied with probabilities of 0.5, 0.05, 0.4, and 0.5, respectively. These steps ensure a robust and diverse training dataset to improve generalization performance.

## 4. Results

In this section, we present experimental results, comparing our method against state-of-the-art (SOTA) techniques on widely-used datasets for the task of Unsupervised Domain Adaptation (UDA) for Object ReID.

*4.1. Dataset description*

We evaluate the effectiveness of our proposal by conducting evaluations on three benchmark datasets: Market-1501 [83], CUHK03 [93], and MSMT17 [39] for Person ReID and two benchmark datasets: Veri-776 [1], VehicleID [84], and VERI-Wild [94] for Vehicle ReID.

**Market-1501** [83] contains 32,668 images of 1,501 individuals captured from six different camera views. The training set includes 12,936 images representing 751 identities, while the testing set comprises 3,368 query images and 19,732 gallery images, covering the remaining 750 identities.

**CUHK03** [93] features 14,097 images of 1,467 unique individuals, recorded by 6 campus cameras, with each identity captured by 2 cameras. The dataset provides two types of annotations: manually labeled bounding boxes and those generated by an automatic detector. For both training and testing, we utilize the manually annotated bounding boxes. Additionally, we follow a more rigorous testing protocol proposed in [95], which splits the dataset into 767 identities (7,365 images) for training and 700 identities for testing, with 5,332 images in the gallery and 1,400 images in the query set.

**MSMT17** [39] is a large-scale dataset comprising 126,441 bounding boxes of 4,101 identities, recorded by 12 outdoor and 3 indoor cameras (15 cameras total) during three periods of the day (morning, afternoon, and noon) over 4 different days. The training set includes 32,621 images featuring 1,041 identities, while the testing set contains 93,820 images representing 3,060 identities. The testing set is further divided into 11,659 query images and 82,161 gallery images. Especially, MSMT17 is significantly larger in scale than both Market-1501 and CUHK03.

**VeRi-776** [1] was collected from 20 real-world surveillance cameras in an urban area under diverse conditions, such as orientations, illuminations, and occlusions. It comprises over 50,000 images of 776 vehicles, and approximately 9000 trajectories. The dataset provides a variety of labels, including identity annotations, vehicle attributes, and spatiotemporal information. It is divided into two subsets for training and testing: the training set contains 37,778 images of 576 vehicles, while the test set consists of 11,579 images of the remaining 200 vehicles.

**VehicleID** [84] contains vehicle images captured by real-world cameras during the daytime. Each subject in the dataset has numerous images taken from the front and back, with some images annotated with model information to aid vehicle identification. The training set comprises 110,178 images of 13,134 vehicles. The test set is divided into three sections: Test800, with 6,532 query images and 800 gallery images of 800 vehicles; Test1600, with 11,385 query images and 1,600 gallery images of 1,600 vehicles; and Test2400, with 17,638 query images and 2,400 gallery images of 2,400 subjects. Following the evaluation protocol of the authors [84], each testing subset divides the query, and gallery sets by randomly selecting one image per subject for the query subset, while the remaining images for each subject form the gallery subset.

**VERI-Wild** [94] is a large-scale dataset comprising 416,314 vehicle images across 40,671 unique identities. These images were collected using a wide-area surveillance system equipped with 174 cameras, spanning an urban region of over 200 km². The camera network operated continuously, capturing vehicle footage 24 hours a day for an entire month. The dataset is split into a training set and three testing subsets. The training set includes 277,797 images of 30,671 vehicle identities. The testing set is further divided into three parts: Test3000 (small) with 41,816 images, Test5000 (medium) with 69,389 images, and Test10000 (large) with 138,517 images.

The comprehensive overview of the datasets utilized in this document is presented in Table 3.

**Table 3.** Details of datasets used in this manuscript.

| Category | Dataset | Cameras | Training Set (ID/Image) | Test Set (ID/Image) | |
|---|---|---|---|---|---|
| | | | | Gallery | Query |
| Person ReID | Market-1501 | 6 | 751/12,936 | 750/19,732 | 750/3,368 |
| | CUHK03 | 2 | 767/7,365 | 700/5,332 | 700/1,400 |
| | MSMT17 | 15 | 1,401/32,621 | 3,060/82,161 | 3,060/11,659 |
| Vehicle ReID | VeRi-776 | 20 | 576/37,778 | 200/11,579 | 200/1,678 |
| | VehicleID | - | 13,134/110,178 | Test800: 800/800 | Test800: 800/6,532 |

|  |  |  | Test1600: 1,600/1,600 | Test1600: 1,600/11,395 |
|---|---|---|---|---|
|  |  |  | Test2400: 2,400/2,400 | Test2400: 2,400/17,638 |
|  |  |  | Test3000: 3,000/38,816 | Test3000: 3,000/3,000 |
| VERI-Wild | 174 | 30,671/277,794 | Test5000: 5,000/64,389 | Test5000: 5,000/5,000 |
|  |  |  | Test10000: 10,000/128,517 | Test10000: 10,000/10,000 |

*4.2. Evaluation metrics*

For the cross-domain Object ReID task, we utilize Rank-$k$ accuracy (where $k \in \{1, 5, 10\}$ and mean average precision (mAP) to evaluate overall performance on test images.

**Rank Ratio Accuracy (Rank-$k$)**: The ranking process involves comparing the features extracted from a query object image $i$ with all images in the gallery. This comparison results in a list of images sorted in descending order of similarity, with the most similar images appearing at the top. According to the ground truth of the selected dataset, the position within this sorted list where an image corresponds to the same object as the query image determines its rank. The Rank-$k$ metric reflects the algorithm's accuracy in correctly identifying object images within the top $k$ ranks among the retrieved results for each query:

$$Rank\text{-}k = \frac{\sum_{i=1}^{M} gt(i,k)}{M} \tag{16}$$

Here, $M$ represents the total number of probe images queried from the gallery, and $gt(i, k)$ is a binary function:

$$gt(i,k) = \begin{cases} 1 \text{ if there are positive samples } i \text{ within the top } n \text{ ranking results} \\ 0 \text{ otherwise} \end{cases}. \tag{17}$$

**Mean Average Precision (mAP)**: In object ReID, where models produce a ranked list of images, it is crucial to consider the position of each image within the list. For each probe image, the average precision (AP) is calculated as follows:

$$AP = \frac{\sum_{j=1}^{N} p(j) \times gt(j)}{N} \tag{18}$$

where $N$ is the total number of images in the gallery set. The values $p(j)$ and $gt(j)$ represent the precision at the $j$-th position in the ranking list and a binary function, respectively. If the probe matches the $j$-th element, then $gt(j) = 1$; otherwise, $gt(j) = 0$. The mean average precision (mAP) across all probe images is then computed using the $AP$ values:

$$mAP = \frac{\sum_{i=1}^{M} AP(i)}{M} \tag{19}$$

Here, $M$ denotes the total number of probe images queried, and $AP(i)$ is the average precision calculated for each probe image $i$.

*4.3. Benchmark on Person ReID*

Our study begins by comparing CORE-ReID V2 with state-of-the-art (SOTA) methods on two domain adaptation tasks: Market → CUHK and CUHK → Market (Table 4). We then expand the evaluation to include two additional tasks: Market → MSMT and CUHK → MSMT (Table 5). In these comparisons, "Baseline" refers to the CORE-ReID method developed in our previous work, while CORE-ReID V2 represents the framework proposed in this paper. The term "Direct Transfer" indicates that the model is trained on the source domain and directly evaluated on the target domain, without applying any pseudo-labeling strategy. Additionally, CORE-ReID V2 Tiny is a lightweight version utilizing the smaller ResNet18 backbone. The evaluation metrics include mAP (%) and rank (R) at $k$ accuracy (%).

**Table 4.** Experimental results of the proposed CORE-ReID V2 framework and SOTA methods (Acc %) on Market-1501 and CUHK03 datasets. **Bold values** represent the best results while Underline values indicate the second-best performance.

|  |  | Market → CUHK | | | | CUHK → Market | | | |
|---|---|---|---|---|---|---|---|---|---|
| **Method** | **Reference** | mAP | R-1 | R-5 | R-10 | mAP | R-1 | R-5 | R-10 |
| SNR [a] [96] | CVPR 2020 | 17.5 | 17.1 | - | - | 52.4 | 77.8 | - | - |
| UDAR [14] | PR 2020 | 20.9 | 20.3 | - | - | 56.6 | 77.1 | - | - |
| QAConv$_{50}$ [a] [97] | ECCV 2020 | 32.9 | 33.3 | - | - | 66.5 | 85.0 | - | - |
| M$^3$L [a] [98] | CVPR 2021 | 35.7 | 36.5 | - | - | 62.4 | 82.7 | - | - |
| MetaBIN [a] [99] | CVPR 2021 | 43.0 | 43.1 | - | - | 67.2 | 84.5 | - | - |
| DFH-Baseline [100] | CVPR 2022 | 10.2 | 11.2 | - | - | 13.2 | 31.1 | - | - |
| DFH [a] [100] | CVPR 2022 | 27.2 | 30.5 | - | - | 31.3 | 56.5 | - | - |
| META [a] [101] | ECCV 2022 | 47.1 | 46.2 | - | - | 76.5 | 90.5 | - | - |
| ACL [a] [102] | ECCV 2022 | 49.4 | 50.1 | - | - | 76.8 | 90.6 | - | - |
| RCFA [103] | Electronics 2023 | 17.7 | 18.5 | 33.6 | 43.4 | 34.5 | 63.3 | 78.8 | 83.9 |
| CRS [104] | JSJTU 2023 | - | - | - | - | 65.3 | 82.5 | 93.0 | 95.9 |
| MTI [105] | JVCIR 2024 | 16.3 | 16.2 | - | - | - | - | - | - |
| PAOA+ [a] [106] | WACV 2024 | 50.3 | 50.9 | - | - | 77.9 | 91.4 | - | - |
| Baseline (CORE-ReID) [11] | Software 2024 | <u>62.9</u> | <u>61.0</u> | <u>79.6</u> | <u>87.2</u> | <u>83.6</u> | <u>93.6</u> | <u>97.3</u> | <u>98.7</u> |
| Direct Transfer | Ours | 23.9 | 24.6 | 40.3 | 48.9 | 35.5 | 63.3 | 77.8 | 83.2 |
| CORE-ReID V2 Tiny (ResNet18) | Ours | 33.0 | 31.9 | 48.9 | 59.1 | 60.3 | 83.4 | 91.8 | 94.7 |
| CORE-ReID V2 | Ours | **66.4** | **66.9** | **83.4** | **88.9** | **84.5** | **93.9** | **97.6** | **98.7** |

**Table 5.** Experimental results of the proposed CORE-ReID framework and SOTA methods (Acc %) from Market-1501 and CUHK03 source datasets to target domain MSMT17 dataset. **Bold values** represent the best results while Underline values indicate the second-best performance. [a] denotes the method uses multiple source datasets, [b] indicates the implementation is based on the author's code.

|  |  | Market → MSMT | | | | CUHK → MSMT | | | |
|---|---|---|---|---|---|---|---|---|---|
| **Method** | **Reference** | mAP | R-1 | R-5 | R-10 | mAP | R-1 | R-5 | R-10 |
| NRMT [107] | ECCV 2020 | 19.8 | 43.7 | 56.5 | 62.2 | - | - | - | - |
| DG-Net++ [87] | ECCV 2020 | 22.1 | 48.4 | - | - | - | - | - | - |
| MMT [15] | ICLR 2020 | 22.9 | 52.5 | - | - | 13.5 [b] | 30.9 [b] | 44.4 [b] | 51.1 [b] |
| UDAR [14] | PR 2020 | 12.0 | 30.5 | - | - | 11.3 | 29.6 | - | - |
| Dual-Refinement [108] | ArXiv 2020 | 25.1 | 53.3 | 66.1 | 71.5 | - | - | - | - |
| SNR [a] [96] | CVPR 2020 | - | - | - | - | 7.7 | 22.0 | - | - |
| QAConv$_{50}$ [a] [97] | ECCV 2020 | - | - | - | - | 17.6 | 46.6 | - | - |
| M$^3$L [a] [98] | CVPR 2021 | - | - | - | - | 17.4 | 38.6 | - | - |
| MetaBIN [a] [99] | CVPR 2021 | - | - | - | - | 18.8 | 41.2 | - | - |
| RDSBN [109] | CVPR 2021 | 30.9 | 61.2 | 73.1 | 77.4 | - | - | - | - |
| ClonedPerson [110] | CVPR 2022 | 14.6 | 41.0 | - | - | 13.4 | 42.3 | - | - |
| META [a] [101] | ECCV 2022 | - | - | - | - | 24.4 | 52.1 | - | - |
| ACL [a] [102] | ECCV 2022 | - | - | - | - | 21.7 | 47.3 | - | - |
| CLM-Net [111] | NCA 2022 | 29.0 | 56.6 | 69.0 | 74.3 | - | - | - | - |
| CRS [104] | JSJTU 2023 | 22.9 | 43.6 | 56.3 | 62.7 | 22.2 | 42.5 | 55.7 | 62.4 |
| HDNet [112] | IJMLC 2023 | 25.9 | 53.4 | 66.4 | 72.1 | - | - | - | - |
| DDNet [113] | AI 2023 | 28.5 | 59.3 | 72.1 | 76.8 | - | - | - | - |
| CaCL [114] | ICCV 2023 | 36.5 | 66.6 | 75.3 | 80.1 | - | - | - | - |
| PAOA+ [a] [106] | WACV 2024 | - | - | - | - | 26.0 | 52.8 | - | - |
| OUDA [115] | WACV 2024 | 20.2 | 46.1 | - | - | - | - | - | - |
| M-BDA [116] | VCIR 2024 | 26.7 | 51.4 | 64.3 | 68.7 | - | - | - | - |

| | | | | | | | | |
|---|---|---|---|---|---|---|---|---|
| UMDA [117] | VCIR 2024 | 32.7 | 62.4 | 72.7 | 78.4 | - | - | - | - |
| Baseline (CORE-ReID) [11] | Software 2024 | <u>41.9</u> | <u>69.5</u> | <u>80.3</u> | <u>84.4</u> | <u>40.4</u> | <u>67.3</u> | <u>79.0</u> | <u>83.1</u> |
| Direct Transfer | Ours | 11.7 | 30.2 | 42.9 | 48.0 | 35.5 | 63.3 | 77.8 | 82.7 |
| CORE-ReID V2 Tiny (ResNet18) | Ours | 35.8 | 64.7 | 76.6 | 80.8 | 18.8 | 44.2 | 57.1 | 62.3 |
| CORE-ReID V2 | Ours | **44.1** | **71.3** | **82.4** | **86.0** | **40.7** | **68.7** | **79.7** | **83.4** |

The results highlight that CORE-ReID V2 significantly outperforms existing SOTA methods, demonstrating the effectiveness of our approach. By incorporating the Ensemble Fusion++ component with ECAB and proposed SECAB, CORE-ReID V2 achieves substantial improvements over the original CORE-ReID. Notably, CORE-ReID V2 surpasses PAOA+ by large margins, achieving mAP improvements of 16.1% and 6.6% on the Market → CUHK and CUHK → Market tasks, respectively, even though PAOA+ utilizes additional training data. Additionally, our framework delivers significant enhancements over CACL and PAOA+, achieving mAP gains of 7.6% and 14.7% mAP on Market → MSMT and CUHK → MSMT tasks, respectively.

*4.4. Benchmark on Vehicle ReID*

We evaluate CORE-ReID V2 against state-of-the-art methods on VehicleID → VeRi-776 (**Table 6**), VehicleID → VERI-Wild (Table 7), and VeRi-776 → VehicleID (Table 8) tasks. "Baseline" refers to the implementation based on CORE-ReID [11] with Ensemble Fusion component, while CORE-ReID V2 is the proposed algorithm. "Direct Transfer" indicates that the model is trained on the source domain and directly evaluated on the target domain, without any clustering-based pseudo-labeling operation. CORE-ReID V2 Tiny is a lightweight variant using ResNet18. Metrics include mAP (%) and rank (R) at *k* accuracy (%).

**Table 6**. Experimental results of CORE-ReID V2 framework and SOTA methods on VehicleID → VeRi-776. **Bold values** represent the best results while <u>Underline values</u> indicate the second-best performance.

| | | VehicleID → VeRi-776 | | | |
|---|---|---|---|---|---|
| **Method** | **Reference** | **mAP** | **R-1** | **R-5** | **R-10** |
| FACT [1] | ECCV 2016 | 18.75 | 52.21 | 72.88 | - |
| PUL [42] | ACM 2018 | 17.06 | 55.24 | 67.34 | - |
| SPGAN [66] | CVPR 2018 | 16.4 | 57.4 | 70.0 | 75.6 |
| VR-PROUD [118] | PR 2019 | 22.75 | 55.78 | 70.02 | - |
| ECN [119] | CVPR 2019 | 20.06 | 57.41 | 70.53 | - |
| MMT [15] | ICLR 2020 | 35.3 | 74.6 | 82.6 | - |
| SPCL [44] | NIPS 2020 | 38.9 | 80.4 | 86.8 | - |
| PAL [120] | IJCAI 2020 | 42.04 | 68.17 | 79.91 | - |
| UDAR [14] | PR 2020 | 35.80 | 76.90 | 85.80 | <u>89.00</u> |
| ML [121] | ICME 2021 | 36.90 | 77.80 | 85.50 | - |
| PLM [122] | Sci.China 2022 | 47.37 | 77.59 | 87.00 | - |
| VDAF [123] | MTA 2023 | 24.86 | 46.32 | 55.17 | - |
| CSP+FCD [124] | Elec 2023 | 45.60 | 74.30 | 83.70 | - |
| MGR-GCL [5] | ArXiv 2024 | 48.73 | <u>79.29</u> | 87.95 | - |
| MATNet+DMDU [125] | ArXiv 2024 | <u>49.25</u> | 79.13 | <u>88.97</u> | - |
| Baseline | Ours | 47.70 | 78.12 | 86.23 | 88.14 |
| Direct Transfer | Ours | 22.71 | 62.04 | 71.79 | 76.32 |
| CORE-ReID V2 Tiny (ResNet18) | Ours | 40.17 | 73.00 | 81.41 | 85.40 |
| CORE-ReID V2 | Ours | **49.50** | **80.15** | **89.05** | **90.29** |

**Table 7.** Experimental results of the proposed CORE-ReID V2 framework and SOTA methods on VehicleID → VERI-Wild. **Bold values** represent the best results while <u>Underline values</u> indicate the second-best performance.

| | | VehicleID → VERI-Wild | | | | | | | | | | | |
|---|---|---|---|---|---|---|---|---|---|---|---|---|---|
| | | Test3000 | | | | Test5000 | | | | Test10000 | | | |
| Method | Reference | mAP | R-1 | R-5 | R-10 | mAP | R-1 | R-5 | R-10 | mAP | R-1 | R-5 | R-10 |
| SPGAN [66] | CVPR 2018 | 24.1 | 59.1 | 76.2 | - | 21.6 | 55.0 | 74.5 | - | 17.5 | 47.4 | 66.1 | - |
| ECN [119] | CVPR 2019 | 34.7 | 73.4 | 88.8 | - | 30.6 | 68.6 | 84.6 | - | 24.7 | 61.0 | 78.2 | - |
| MMT [15] | ICLR 2020 | 27.7 | 55.6 | 77.4 | - | 23.6 | 47.7 | 71.5 | - | 18.0 | 40.2 | 65.0 | - |
| SPCL [44] | NIPS 2020 | 25.1 | 48.8 | 72.8 | - | 21.5 | 42.0 | 66.1 | - | 16.6 | 32.7 | 55.7 | - |
| UDAR [14] | PR 2020 | 30.0 | 68.4 | 85.3 | - | 26.2 | 62.5 | 81.8 | - | 20.8 | 53.7 | 73.9 | - |
| AE [126] | CCA 2020 | 29.9 | 67.0 | 68.5 | - | 26.2 | 61.8 | 81.5 | - | 20.9 | 53.1 | 73.7 | - |
| DLVL [18] | Elec 2024 | 31.4 | 59.9 | 80.7 | - | 27.3 | 51.9 | 74.9 | - | 21.7 | 41.8 | 65.8 | - |
| Baseline | Ours | <u>39.8</u> | <u>75.2</u> | <u>89.3</u> | <u>91.6</u> | <u>34.5</u> | <u>69.6</u> | <u>81.7</u> | <u>88.7</u> | <u>26.8</u> | <u>61.1</u> | <u>79.6</u> | <u>81.3</u> |
| Direct Transfer | Ours | 20.9 | 48.2 | 64.3 | 70.7 | 18.9 | 44.3 | 60.9 | 66.9 | 15.6 | 38.0 | 53.3 | 59.8 |
| CORE-ReID V2 Tiny (ResNet18) | Ours | 28.6 | 56.5 | 74.9 | 80.2 | 23.1 | 52.1 | 70.6 | 78.4 | 19.9 | 48.1 | 66.3 | 74.6 |
| CORE-ReID V2 | Ours | **40.2** | **76.6** | **90.2** | **92.1** | **34.9** | **70.2** | **86.2** | **89.3** | **27.8** | **62.1** | **79.8** | **82.3** |

**Table 8.** Experimental results of the proposed CORE-ReID V2 framework and SOTA methods on VeRi-776 → VehicleID. **Bold values** represent the best results while <u>Underline values</u> indicate the second-best performance.

| | | VeRi-776 → VehicleID | | | | VeRi-776 → VehicleID | | | | VeRi-776 → VehicleID | | | |
|---|---|---|---|---|---|---|---|---|---|---|---|---|---|
| | | Test800 | | | | Test1600 | | | | Test2400 | | | |
| Method | Reference | mAP | R-1 | R-5 | R-10 | mAP | R-1 | R-5 | R-10 | mAP | R-1 | R-5 | R-10 |
| FACT [1] | ECCV 2016 | - | 49.53 | 67.96 | - | - | 44.63 | 64.19 | - | - | 39.91 | 60.49 | - |
| Mixed Diff+CCL [84] | CVPR 2016 | - | 49.00 | 73.50 | - | - | 42.80 | 66.80 | - | - | 38.20 | 61.60 | - |
| PUL [42] | ACM 2018 | 43.90 | 40.03 | 56.03 | - | 37.68 | 33.83 | 49.72 | - | 34.71 | 30.90 | 47.18 | - |
| PAL [120] | IJCAI 2020 | 53.50 | 50.25 | 64.91 | - | 48.05 | 44.25 | 60.95 | - | 45.14 | 41.08 | 59.12 | - |
| UDAR [14] | PR 2020 | 59.60 | 54.00 | 66.10 | 72.01 | 55.30 | 48.10 | 64.10 | 70.20 | 52.90 | 45.20 | 62.60 | 69.14 |
| ML [121] | ICME 2021 | 61.60 | 54.80 | 69.20 | - | 48.70 | 40.30 | 57.70 | - | 45.00 | 36.50 | 54.10 | - |
| PLM [122] | Sci.China 2022 | 54.85 | 51.23 | 67.11 | - | 49.41 | 45.40 | 63.37 | - | 46.00 | 41.73 | 60.94 | - |
| CSP+FCD [124] | Elec 2023 | 51.90 | 54.40 | 67.40 | - | 46.50 | 52.70 | 65.60 | - | 42.70 | 45.90 | 60.30 | - |
| VDAF [123] | MTA 2023 | - | - | - | - | - | 47.03 | 64.86 | - | - | 43.69 | 61.76 | - |
| MGR-GCL [5] | ArXiv 2024 | 55.24 | 52.38 | <u>75.29</u> | - | 50.56 | 45.88 | 67.65 | - | 47.59 | 42.83 | 64.36 | - |
| DMDU [125] | TITS 2024 | 61.83 | 55.61 | 68.25 | - | 56.73 | <u>53.28</u> | 63.56 | - | 53.97 | 47.59 | 61.85 | - |
| Baseline | Ours | <u>64.28</u> | <u>56.16</u> | 74.55 | <u>81.15</u> | <u>60.02</u> | 51.84 | <u>71.62</u> | <u>78.08</u> | <u>56.15</u> | <u>47.85</u> | <u>66.89</u> | <u>75.27</u> |
| Direct Transfer | Ours | 61.28 | 53.50 | 69.81 | 76.13 | 57.23 | 48.57 | 67.05 | 73.77 | 52.31 | 44.04 | 61.08 | 68.60 |
| CORE-ReID V2 Tiny (ResNet18) | Ours | 63.87 | 55.18 | 73.43 | 81.11 | 59.69 | 50.05 | 70.88 | 77.75 | 55.14 | 45.99 | 65.07 | 73.54 |
| CORE-ReID V2 | Ours | **67.04** | **58.32** | **76.51** | **84.32** | **63.02** | **53.49** | **74.36** | **81.85** | **57.99** | **48.62** | **68.30** | **77.11** |

Across multiple domain adaptation scenarios (VeRi-776 → VehicleID, VehicleID → VERI-Wild, and VehicleID → VeRi-776), the proposed CORE-ReID V2 framework demonstrates superior performance compared to existing state-of-the-art (SOTA) methods. Specifically, our evaluation includes supervised approaches such as FACT and Mixed Diff+CCL, as well as unsupervised Person ReID methods PUL and UDAR. Additionally, we incorporate leading SOTA techniques, including PAL, MMT, SPCL PLM, CSP+FCD, and DMDU, for a comprehensively comparative analysis.

In the VeRi-776 → VehicleID adaptation setting, CORE-ReID V2 achieves 67.04%, 63.02%, and 57.99% mAP across three test modes, surpassing the DMDU method by 5.21%, 6.29%, and 4.02% in each case. For the VehicleID → VERI-Wild scheme, our framework records mAP of 40.2%, 34.9% and 27.8%, outperforming ECN method by 5.5%, 4.3%, and 3.1% on the Test3000, Test5000 and Test10000 evaluation mode, respectively. In the VehicleID → VeRi-776 scenario, MGR-GCL and MATNet+DMDU attain 48.73% and 49.25% mAP, respectively. CORE-ReID V2 outperforms all competing methods, achieving a mAP of 49.50% and Rank-1 accuracy of 80.15%, setting its position as a new SOTA approach.

*4.5 Ablation Study*

**Feature Maps Visualization:** To validate our approach, we employ Grad-CAM [127] to visualize feature maps at the global feature level. Key features for each person and vehicle are highlighted using heatmaps, where color intensity indicates importance - blue represents less significant regions, while red denotes the most crucial areas for Object Re-identification. As illustrated in Figure 10 and Figure 11, the essential features are concentrated on the target person's body and the vehicle's structure. Furthermore, the heatmaps exhibit similar distributions between the original and flipped images, reinforcing the performance of our method. This consistency aligns with the accuracy results reported above, further validating the effectiveness of CORE-ReID V2.

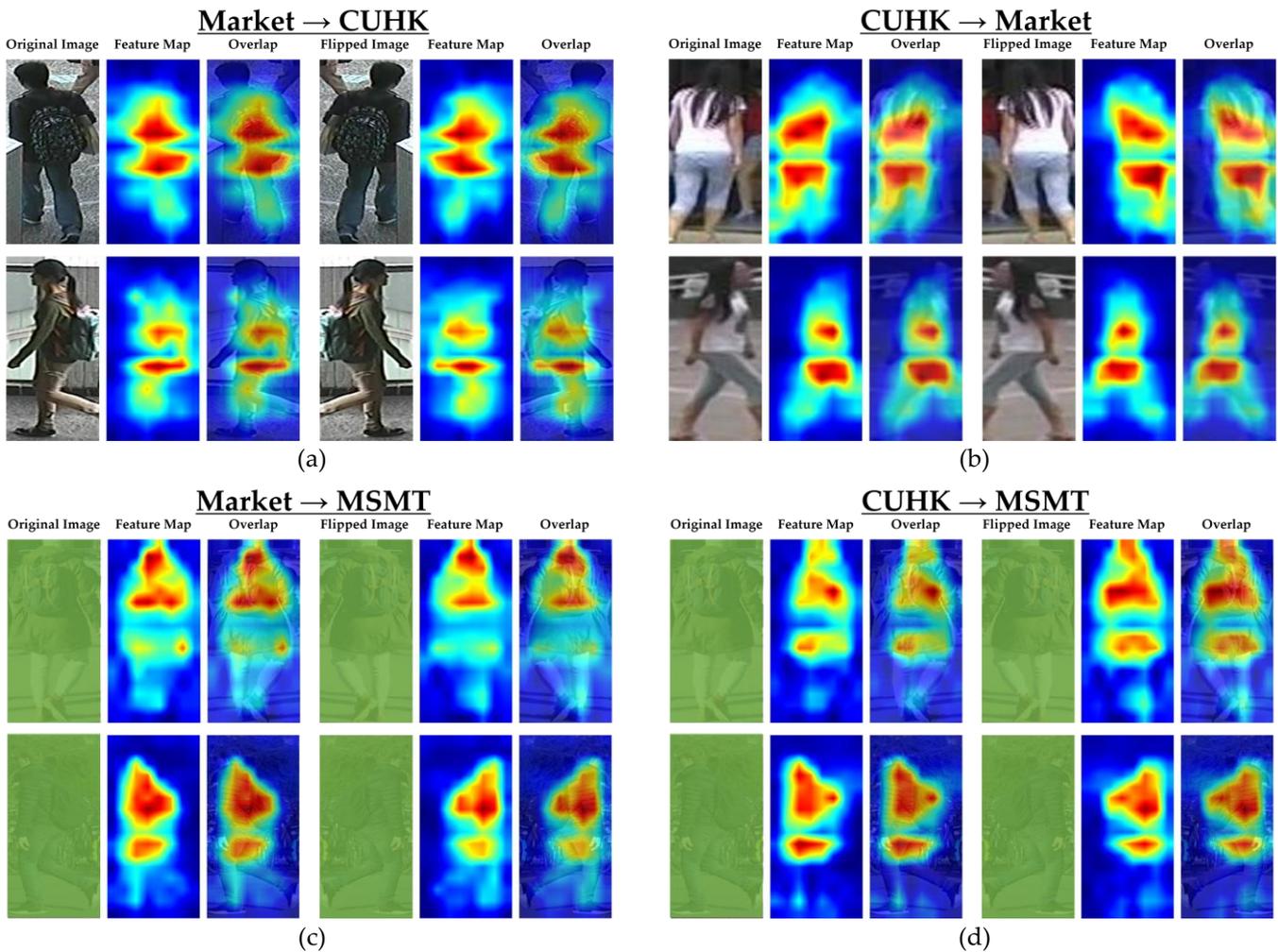

**Figure 10.** Feature maps visualization using Grad-CAM [127]. (a), (b), (c), and (d) illustrate the feature maps of those pairs on Market→CUHK, CUHK→Market, Market→MSMT, and CUHK→MSMT, respectively.

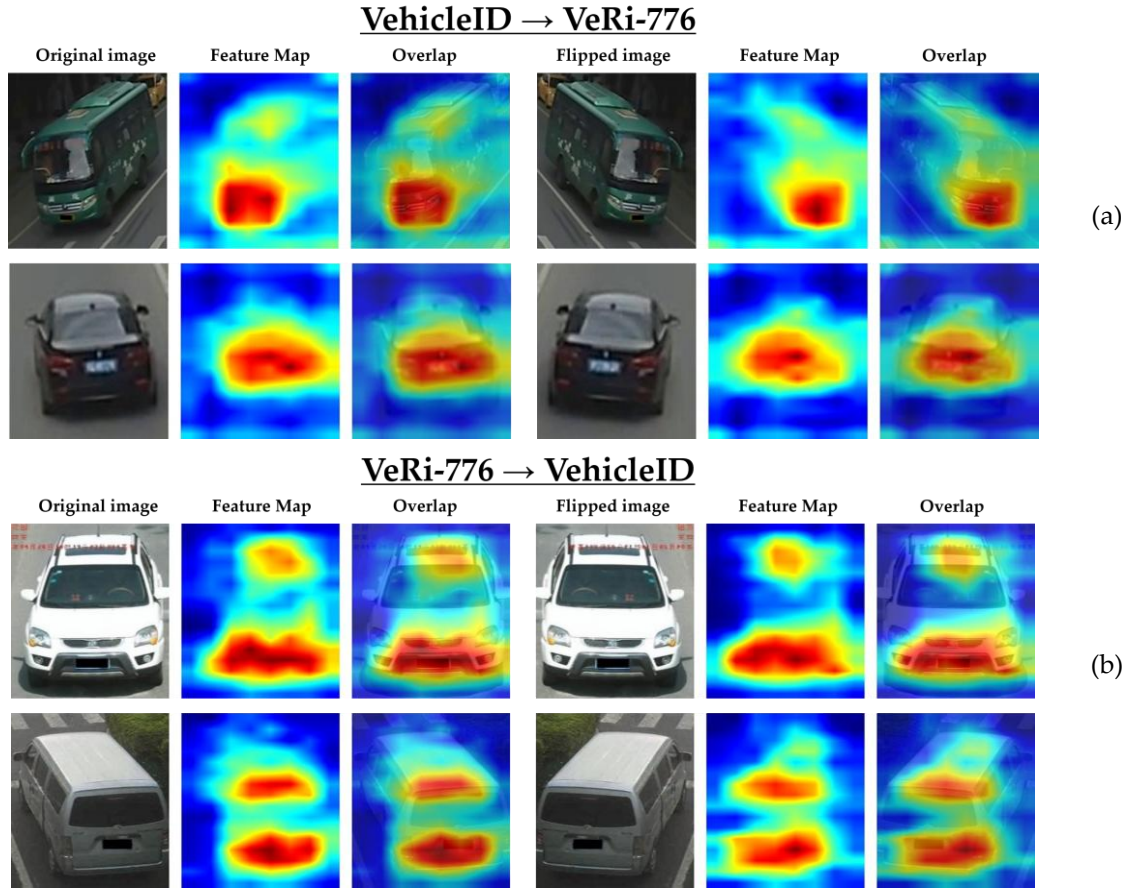

**Figure 11.** Feature maps visualization using Grad-CAM [127]. (a), (b) illustrate the feature maps of those pairs on VehicleID→ VeRi-776 and VeRi-776 → VehicleID respectively.

In the Market → MSMT and CUHK → MSMT scenarios, the Market → MSMT model demonstrates a slightly superior ability to extract important features. The heatmaps reveal a more concentrated distribution in the middle and lower body regions for both the original and flipped images. This observation may explain the higher accuracy achieved by the Market → MSMT model compared to the CUHK → MSMT model, as reported in Table 5.

**K-means Clustering Settings:** we utilize the K-Means clustering approach to generate pseudo-labels for the target domain, with parameters varying across different datasets. As shown in Table 9, our framework achieves optimal performance on Market → CUHK, CUHK → Market, Market → MSMT, CUHK → MSMT, VehicleID → VeRi-776, and VeRi-776 → VehicleID Small with cluster settings of 900, 900, 2000, 2000, 500 and 700, respectively.

**Table 9.** Experimental results on different settings of number of pseudo identities in K-means clustering algorithm for both Person and Vehicle ReID tasks. **Bold values** represent the best results.

| Person ReID | Market → CUHK | | | | CUHK → Market | | | |
|---|---|---|---|---|---|---|---|---|
| Number of Clusters | mAP | R-1 | R-5 | R-10 | mAP | R-1 | R-5 | R-10 |
| Ours ($M_{T,j} = 500$) | 44.4 | 43.2 | 65.3 | 76.4 | 69.4 | 86.8 | 94.9 | 96.7 |
| Ours ($M_{T,j} = 700$) | 57.8 | 59.1 | 76.1 | 83.6 | 81.7 | 92.7 | 97.1 | 98.1 |
| Ours ($M_{T,j} = 900$) | **66.4** | **66.9** | **83.4** | **88.9** | **84.5** | **93.9** | **97.6** | **98.7** |
| **Person ReID** | Market → MSMT | | | | CUHK → MSMT | | | |
| Number of Clusters | mAP | R-1 | R-5 | R-10 | mAP | R-1 | R-5 | R-10 |
| Ours ($M_{T,j} = 2000$) | **44.1** | **71.3** | **82.4** | **86.0** | **40.68** | **68.66** | **79.74** | **83.36** |
| Ours ($M_{T,j} = 2500$) | 41.1 | 68.9 | 80.5 | 84.2 | 38.91 | 67.26 | 78.97 | 82.80 |
| Ours ($M_{T,j} = 3000$) | 38.9 | 67.2 | 79.0 | 83.2 | 35.8 | 64.7 | 76.6 | 80.8 |
| Vechile ReID | VehicleID → VeRi-776 | | | | VeRi-776 → VehicleID Small | | | |
| Number of Clusters | mAP | R-1 | R-5 | R-10 | mAP | R-1 | R-5 | R-10 |
| Ours ($M_{T,j} = 500$) | 49.50 | **80.15** | **89.05** | **90.29** | 66.60 | 58.20 | 75.90 | 83.70 |
| Ours ($M_{T,j} = 700$) | **49.63** | 79.14 | 86.65 | 89.69 | **67.04** | **58.32** | **76.51** | **84.32** |
| Ours ($M_{T,j} = 900$) | 48.61 | 79.02 | 86.29 | 89.15 | 66.70 | 57.50 | 77.60 | 84.20 |

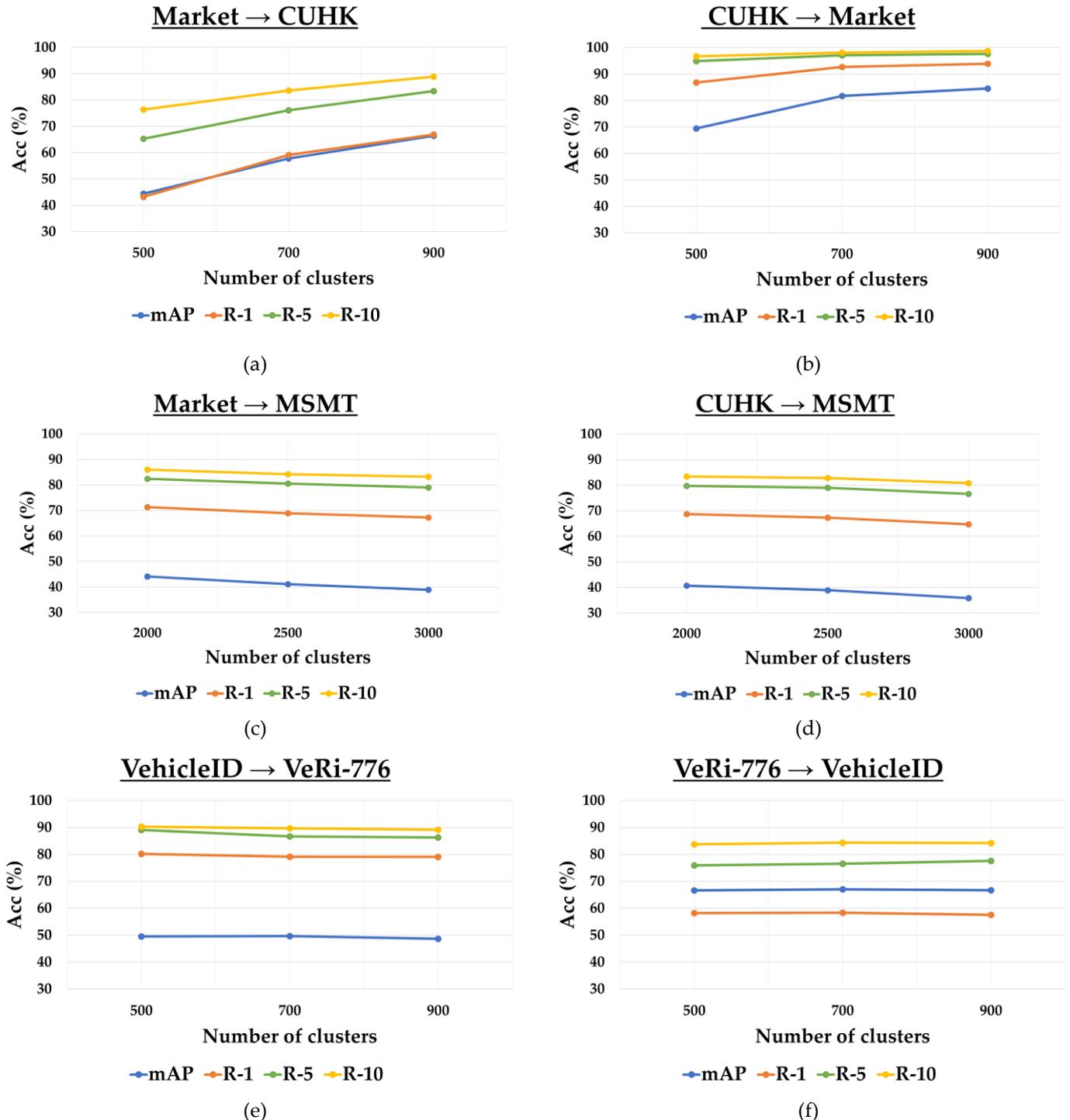

**Figure 12.** Impact of clustering parameter $M_{T,j}$. Results on (a) Market → CUHK, (b) CUHK → Market, (c) Market → MSMT, (d) CUHK → MSMT, (e) VehicleID → VeRi-776, and (f) VeRi-776 → VehicleID Small.

Figure 12 shows that the performance of our approach varies depending on the dataset pairs and the clustering parameter values ($M_{T,j}$) utilized.

**Greedy K-means++ Initialization:** we enhance clustering performance by employing the greedy K-means++ initialization strategy, which optimally balances randomness and centroid selection to improve cluster quality. This approach not only strengthens feature learning but also ensures a more stable pseudo-label generation, addressing challenges associated with obscure learning. Table 10 presents the experimental results comparing greedy K-means++ initialization with a random approach.

**Table 10.** Experimental results using the greedy K-means++ initialization and random approach. The clustering parameter values $M_{T,j}$ are carried out from the study of K-means clustering settings. **Bold values** represent better results.

| Person ReID | Market → CUHK ($M_{T,j} = 900$) | | | | CUHK → Market ($M_{T,j} = 900$) | | | |
|---|---|---|---|---|---|---|---|---|
| Method | mAP | R-1 | R-5 | R-10 | mAP | R-1 | R-5 | R-10 |
| Ours (Random) | 63.6 | 63.8 | 80.9 | 87.8 | 83.8 | 93.6 | 97.4 | 98.6 |
| Ours (Greedy Initialization) | **66.4** | **66.9** | **83.4** | **88.9** | **84.5** | **93.9** | **97.6** | **98.7** |
| **Person ReID** | Market → MSMT ($M_{T,j} = 2000$) | | | | CUHK → MSMT ($M_{T,j} = 2000$) | | | |
| Method | mAP | R-1 | R-5 | R-10 | mAP | R-1 | R-5 | R-10 |
| Ours (Random) | 42.2 | 69.7 | 80.2 | 84.9 | 40.5 | 67.6 | 78.8 | 83.1 |
| Ours (Greedy Initialization) | **44.1** | **71.3** | **82.4** | **86.0** | **40.7** | **68.7** | **79.7** | **83.4** |
| **Vehicle ReID** | VehicleID → VeRi-776 ($M_{T,j} = 500$) | | | | VeRi-776 → VehicleID Small ($M_{T,j} = 700$) | | | |
| Method | mAP | R-1 | R-5 | R-10 | mAP | R-1 | R-5 | R-10 |
| Ours (Random) | 47.72 | 78.23 | 86.56 | 88.26 | 65.79 | 56.14 | 75.95 | 83.56 |
| Ours (Greedy Initialization) | **49.50** | **80.15** | **89.05** | **90.29** | **67.04** | **58.32** | **76.51** | **84.32** |

**SECAB Configuration:** We use SECAB to enhance global features, leveraging attention mechanisms to emphasize crucial features while suppressing unnecessary ones. To validate the effectiveness of SECAB, we conduct an experiment by removing it from our network, as shown in Table 11.

**Table 11.** Experimental results validating the effectiveness of SECAB in our proposed framework. The clustering parameter values ($M_{T,j}$) are derived from the study of K-means clustering settings. **Bold values** represent better results.

| Person ReID | Market → CUHK ($M_{T,j} = 900$) | | | | CUHK → Market ($M_{T,j} = 900$) | | | |
|---|---|---|---|---|---|---|---|---|
| Method | mAP | R-1 | R-5 | R-10 | mAP | R-1 | R-5 | R-10 |
| Ours (without SECAB) | 65.0 | 65.1 | 82.6 | 87.6 | 83.9 | 93.7 | 97.4 | 98.6 |
| Ours (with SECAB) | **66.4** | **66.9** | **83.4** | **88.9** | **84.5** | **93.9** | **97.6** | **98.7** |
| **Person ReID** | Market → MSMT ($M_{T,j} = 2000$) | | | | CUHK → MSMT ($M_{T,j} = 2000$) | | | |
| Method | mAP | R-1 | R-5 | R-10 | mAP | R-1 | R-5 | R-10 |
| Ours (without SECAB) | 43.2 | 70.3 | 81.8 | 85.2 | 40.5 | 68.0 | 79.2 | 83.1 |
| Ours (with SECAB) | **44.1** | **71.3** | **82.4** | **86.0** | **40.7** | **68.7** | **79.7** | **83.4** |
| **Vehicle ReID** | VehicleID → VeRi-776 ($M_{T,j} = 500$) | | | | VeRi-776 → VehicleID Small ($M_{T,j} = 700$) | | | |
| Method | mAP | R-1 | R-5 | R-10 | mAP | R-1 | R-5 | R-10 |
| Ours (without SECAB) | 48.03 | 78.92 | 87.61 | 88.93 | 65.14 | 57.02 | 75.56 | 82.97 |
| Ours (with SECAB) | **49.50** | **80.15** | **89.05** | **90.29** | **67.04** | **58.32** | **76.51** | **84.32** |

**Ensemble Fusion++ Configuration:** We apply ECAB to refine local features by capturing rich inter-channel dependencies, and SECAB to enhance global features. This design allows for adaptive feature recalibration at both local and global levels. To assess the effectiveness of this configuration, we conducted a series of experiments with different attention settings, as summarized in Table 12, demonstrating that this combination yields the most consistent performance improvement.

**Table 12**. Experimental results with different attention settings in Ensemble Fusion++ component. The clustering parameter values ($M_{T,j}$) are derived from the study of K-means clustering settings. **Bold values** represent the best results.

| Person ReID | Market → CUHK ($M_{T,j} = 900$) | | | | CUHK → Market ($M_{T,j} = 900$) | | | |
|---|---|---|---|---|---|---|---|---|
| Method | mAP | R-1 | R-5 | R-10 | mAP | R-1 | R-5 | R-10 |
| Ours (Top-local: ECAB, Bottom-Local: ECAB, Global: ECAB) | 64.3 | 64.9 | 81.6 | 84.3 | 83.2 | 92.6 | 97.3 | 98.6 |
| Ours (Top-local: SECAB, Bottom-Local: SECAB, Global: SECAB) | 62.3 | 63.2 | 80.2 | 82.8 | 81.0 | 89.6 | 94.5 | 95.1 |
| **Ours (Top-local: ECAB, Bottom-Local: ECAB, Global: SECAB)** | **66.4** | **66.9** | **83.4** | **88.9** | **84.5** | **93.9** | **97.6** | **98.7** |

**Backbone Settings:** we assess the performance of both complex backbones (ResNet50, ResNet101, and ResNet152) and lightweight backbones (ResNet18 and ResNet34) for unsupervised domain adaptation in Object ReID. Through extensive experiments and analysis, we gain insights into the impact of backbone architecture on overall performance, as well as its computational efficiency and suitability for resource-constrained environments, as shown in Table 13. Among the tested configurations, ResNet101 delivers the best performance in Market → CUHK, CUHK → Market, VehicleID → VeRi-776, and VeRi-776 → VehicleID scenarios. All experiments were conducted on two machines, each equipped with dual Quadro RTX 8000 GPUs.

**Table 13.** Experimental results on different settings of ResNet backbones in Market → CUHK, CUHK → Market scenarios. **Bold values** represent the best results.

| Person ReID | Market → CUHK ($M_{T,j} = 900$) | | | | CUHK → Market ($M_{T,j} = 900$) | | | |
|---|---|---|---|---|---|---|---|---|
| Method | mAP | R-1 | R-5 | R-10 | mAP | R-1 | R-5 | R-10 |
| Ours (ResNet18) | 33.0 | 31.9 | 48.9 | 59.1 | 60.3 | 83.4 | 91.8 | 94.7 |
| Ours (ResNet34) | 38.8 | 38.4 | 55.9 | 64.7 | 64.4 | 85.9 | 93.7 | 95.4 |
| Ours (ResNet50) | 64.9 | 64.1 | 81.3 | 87.9 | 83.7 | 93.8 | 97.6 | 98.5 |
| Ours (ResNet101) | **66.4** | **66.9** | **83.4** | **88.9** | **84.5** | **93.9** | **97.6** | **98.7** |
| Ours (ResNet152) | 65.2 | 65.1 | 82.1 | 87.9 | 83.5 | 93.2 | 97.5 | 98.1 |
| **Vehicle ReID** | VehicleID → VeRi-776 ($M_{T,j} = 500$) | | | | VeRi-776 → VehicleID Small ($M_{T,j} = 700$) | | | |
| Method | mAP | R-1 | R-5 | R-10 | mAP | R-1 | R-5 | R-10 |
| Ours (ResNet18) | 40.17 | 73.00 | 81.41 | 85.40 | 63.87 | 55.18 | 73.43 | 81.11 |
| Ours (ResNet34) | 46.62 | 75.92 | 83.73 | 87.49 | 63.80 | 54.80 | 73.60 | 80.30 |
| Ours (ResNet50) | 48.11 | 78.84 | 86.71 | 89.81 | 67.02 | 58.30 | **77.00** | 83.90 |
| Ours (ResNet101) | **49.50** | **80.15** | **89.05** | **90.29** | **67.04** | **58.32** | 76.51 | **84.32** |
| Ours (ResNet152) | 48.07 | 78.26 | 86.73 | 89.96 | 66.97 | 58.23 | 76.49 | 83.86 |

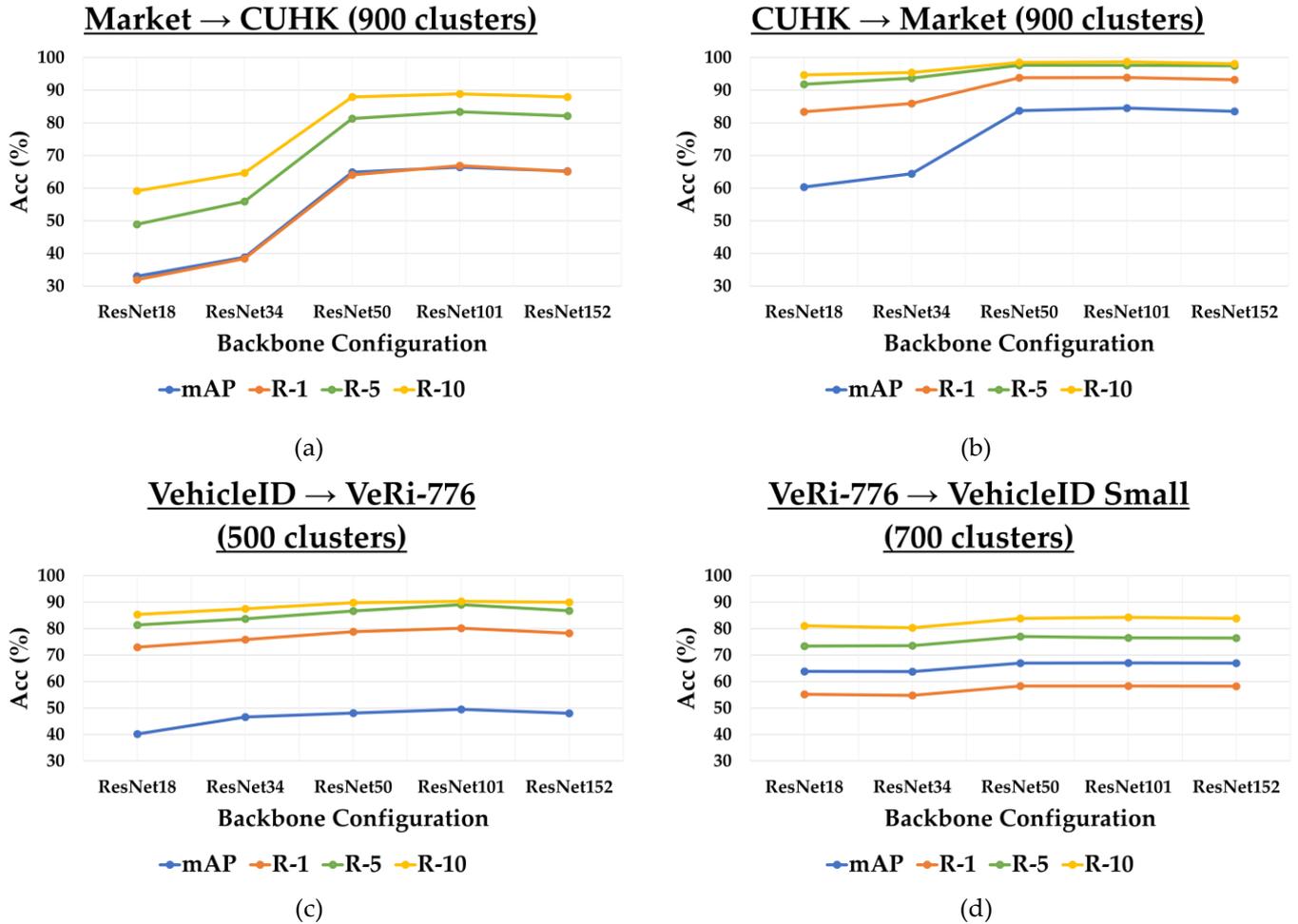

**Figure 13.** Impact of the backbone settings. Results on (a) Market → CUHK, (b) CUHK → Market, (c) VehicleID → VeRi-776, and (d) VeRi-776 → VehicleID Small show that ResNet101 backbone gives the best overall results.

Figure 13 demonstrates that lightweight backbones, such as ResNet18 and ResNet34, are also supported, highlighting the adaptability of our framework to various backbone architectures.

**Computational Complexity Analysis:** One of the key strengths of CORE-ReID V2 lies in its flexibility to support lightweight backbone networks, such as ResNet-18 and ResNet-34, in addition to deeper ones like ResNet-50, 101, and 152. This flexibility allows users to balance accuracy and computational cost, enabling deployment in real-time or edge-based environments with limited resources. To highlight this advantage, we include a comparative table of commonly used ResNet backbones in terms of parameters and Giga Floating-Point Operations per Second (GFLOPs):

**Table 14.** Evaluated using single-image inference on a Quadro RTX 8000 GPU with a batch size of 1, CORE-ReID V2 with ResNet18 and ResNet34 demonstrates significantly lower computational cost (fewer GFLOPs and parameters), and faster processing speed (higher Frames per Second) compared to deeper models.

| CORE-ReID V2 with Backbone | Parameters (millions) | GFLOPs (per image) | Image Size | FPS (using 1 Quadro RTX 8000 GPU) |
|---|---|---|---|---|
| ResNet-18 | 12.97 M | 1.18 | 128*256 | 254 |
| ResNet-34 | 23.08 M | 2.35 | 128*256 | 185 |
| ResNet-50 | 46.62 M | 5.10 | 128*256 | 144 |
| ResNet-101 | 65.61 M | 7.58 | 128*256 | 87 |
| ResNet-152 | 81.26 M | 10.61 | 128*256 | 61 |

By allowing the use of ResNet-18/34 as backbones, CORE-ReID V2 can achieve substantial reductions in training/inference cost, enabling faster deployment while maintaining competitive UDA performance. Additionally, only the teacher model is used during inference, further reducing runtime complexity. We believe these design choices make CORE-ReID V2 an efficient yet effective UDA solution suitable for practical deployment, especially in low-power or embedded AI systems.

## 5. Conclusions

In this paper, we introduced CORE-ReID V2, an enhanced framework designed to address limitations in its predecessor, CORE-ReID, while extending its applicability to Object ReID tasks. By incorporating the novel Ensemble Fusion++ module, which adaptively enhances both local and global features, and utilizing advanced clustering techniques such as greedy KMeans++ initialization, CORE-ReID V2 achieves superior performance in Unsupervised Domain Adaptation (UDA) for Person and Vehicle ReID tasks. Furthermore, support for lightweight backbones like ResNet18 and ResNet34 makes the framework suitable for real-time and resource-constrained applications. Experimental results on widely used UDA Person ReID and Vehicle ReID datasets demonstrate that CORE-ReID V2 outperforms state-of-the-art methods, showcasing its strength and adaptability. These contributions not only push the boundaries of UDA-based Object ReID but also provide a solid foundation for further exploration in this domain.

Despite its advancements, CORE-ReID V2 has several limitations that warrant attention. The scalability of the framework to other datasets such as BV-Person [128], ENTIReID [129], VRID-1 [130], VRAI [131], Vehicle-Rear [132] and V2I-CARLA [133] remains unexplored, which could introduce challenges in both performance and efficiency. Furthermore, the framework's primary focus on Person and Vehicle ReID tasks also limits its exploration of broader Object ReID applications, such as animal or product identification. Moreover, its reliance on the quality of pseudo-labels makes it vulnerable to performance degradation in noisy or highly complex scenarios.

To address these limitations, future work will focus on the following directions. First, to enhance scalability, we plan to evaluate and optimize CORE-ReID V2 across a broader range of Object ReID datasets, including mentioned datasets above. This will allow us to assess the generalization capability of the framework and adapt it to more diverse environments and conditions. Domain-specific normalization strategies and adaptive sampling techniques may be introduced to maintain performance consistency across different dataset characteristics. Second, although this work focuses on person and vehicle Re-ID, the proposed framework is general and can be extended to other object-level UDA tasks. The core mechanisms, such as feature aggregation, pseudo-label refinement, and cross-domain consistency learning, are not restricted to a particular object category. We plan to further evaluate CORE-ReID V2 on general object, such as animal, product, or scene-specific ReID, by incorporating domain-specific priors and augmentation strategies to demonstrate this broader applicability. Third, exploring advanced techniques, such as contrastive learning and adversarial regularization, to mitigate the impact of noisy pseudo-labels and improve model performance. Hopefully, these directions aim to unlock the full potential of CORE-ReID V2 and inspire future advancements in UDA for Object ReID.

**Author Contributions:** Writing—original draft, T.Q.N. and O.D.A.P.; conceptualization, T.Q.N., O.D.A.P. and S.A.I; methodology, T.Q.N. and O.D.A.P.; software, T.Q.N., O.D.A.P. and S.A.I.; validation, T.Q.N., H.D.P. and R.T.; formal analysis, T.Q.N. and O.D.A.P.; investigation, T.Q.N., O.D.A.P. and S.A.I.; resources, T.Q.N., O.D.A.P., H.D.P. and R.T.; data curation, T.Q.N., S.A.I., H.D.P. and R.T.; writing—review and editing, T.Q.N., O.D.A.P., S.A.I., H.D.P. and R.T.; visualization, T.Q.N., H.D.P. and R.T.; supervision, O.D.A.P.; project administration, T.Q.N.; funding

34acquisition, T.Q.N. and O.D.A.P. All authors have read and agreed to the published version of the manuscript.

**Funding:** This research was supported by a Grant-in-Aid for Scientific Research (KAKENHI) from the Japan Society for the Promotion of Science (JSPS). Grant number: 25K15160.

**Institutional Review Board Statement:** Not applicable.

**Informed Consent Statement**: Not applicable.

**Data Availability Statement:** The data (code and models) presented in the study are openly available on GitHub at https://github.com/TrinhQuocNguyen/CORE-ReID-V2 (accessed on 17 January 2025).

**Acknowledgments:** The authors sincerely appreciate the support of CyberCore Co., Ltd. for providing the necessary training environments. They also extend their gratitude to the anonymous reviewers for their insightful feedback, which has contributed to enhancing the quality of this paper. Additionally, the authors would like to thank Raymond Swannack of the School of Information and Computer Science, Iwate Prefectural University, for his assistance in proofreading the manuscript.

**Conflicts of Interest:** Authors Trinh Quoc Nguyen and Syahid Al Irfan were employed by the company CyberCore Co., Ltd. The remaining authors declare that the research was conducted in the absence of any commercial or financial relationships that could be construed as a potential conflict of interest.
**Abbreviations**

The following abbreviations are used in this manuscript:

| | |
|---|---|
| ECAB | Efficient Channel Attention Block |
| BMFN | Bidirectional Mean Feature Normalization |
| CBAM | Convolutional Block Attention Module |
| CNN | Convolutional Neural Network |
| CORE-ReID | Comprehensive Optimization and Refinement through Ensemble fusion in Domain Adaptation for Person Re-identification |
| HHL | Hetero-Homogeneous Learning |
| MMFA | Multi-task Mid-level Feature Alignment |
| MMT | Mutual Mean-Teaching |
| Object ReID | Object Re-identification |
| SECAB | Simplified Efficient Channel Attention Block |
| SOTA | State-Of-The-Art |
| SSG | Self-Similarity Grouping |
| UDA | Unsupervised Domain Adaptation |
| UNRN | Uncertainty-Guided Noise-Resilient Network |